# Improving Statistical Machine Translation
# for a Resource-Poor Language
# Using Related Resource-Rich Languages

**Preslav Nakov**                                                    PNAKOV@QF.ORG.QA
*Qatar Computing Research Institute*
*Qatar Foundation*
*Tornado Tower, Floor 10, P.O. Box 5825*
*Doha, Qatar*

**Hwee Tou Ng**                                                    NGHT@COMP.NUS.EDU.SG
*Department of Computer Science*
*National University of Singapore*
*13 Computing Drive*
*Singapore 117417*

## Abstract

We propose a novel language-independent approach for improving machine translation for resource-poor languages by exploiting their similarity to resource-rich ones. More precisely, we improve the translation from a resource-poor source language $X_1$ into a resource-rich language $Y$ given a bi-text containing a limited number of parallel sentences for $X_1$-$Y$ and a larger bi-text for $X_2$-$Y$ for some resource-rich language $X_2$ that is closely related to $X_1$. This is achieved by taking advantage of the opportunities that vocabulary overlap and similarities between the languages $X_1$ and $X_2$ in spelling, word order, and syntax offer: (1) we improve the word alignments for the resource-poor language, (2) we further augment it with additional translation options, and (3) we take care of potential spelling differences through appropriate transliteration. The evaluation for Indonesian→English using Malay and for Spanish→English using Portuguese and pretending Spanish is resource-poor shows an absolute gain of up to 1.35 and 3.37 BLEU points, respectively, which is an improvement over the best rivaling approaches, while using much less additional data. Overall, our method cuts the amount of necessary "real" training data by a factor of 2–5.

## 1. Introduction

Recent developments in statistical machine translation (SMT), e.g., the availability of efficient implementations of integrated open-source toolkits like Moses (Koehn, Hoang, Birch, Callison-Burch, Federico, Bertoldi, Cowan, Shen, Moran, Zens, Dyer, Bojar, Constantin, & Herbst, 2007), have made it possible to build a prototype system with decent translation quality for any language pair in a few days or even hours. This is so in theory. In practice, doing so requires having a large set of parallel sentence-aligned texts in two languages (*bi-texts*) for that language pair. Such large high-quality bi-texts are rare; except for Arabic, Chinese, and some official languages of the European Union (EU), most of the 6,500+ world languages remain resource-poor from an SMT viewpoint.

The number of resource poor languages becomes even more striking if we consider language *pairs* instead of individual languages. Moreover, even resource-rich language pairs could be poor in bi-texts for a specific domain, e.g., biomedical.

While manually creating a small bi-text could be relatively easy, building a *large* one is hard and time-consuming. Thus, most publicly available bi-texts for SMT come from parliament debates and legislation of multi-lingual countries (e.g., French-English from Canada, and Chinese-English from Hong Kong), or from international organizations like the United Nations and the European Union. For example, the *Europarl* corpus of parliament proceedings consists of about 1.3M parallel sentences (up to 44M words) per language for 11 languages (Koehn, 2005), and the *JRC-Acquis* corpus provides a comparable amount of European legislation in 22 languages (Steinberger, Pouliquen, Widiger, Ignat, Erjavec, Tufis, & Varga, 2006).

Due to the increasing volume of EU parliament debates and the ever-growing European legislation, the official languages of the EU are especially privileged from an SMT perspective. While this includes "classic SMT languages" such as English and French (which were already resource-rich), and some important international ones like Spanish and Portuguese, many of the rest have a limited number of speakers and were resource-poor until a few years ago. Thus, becoming an official language of the EU has turned out to be an easy recipe for getting resource-rich in bi-texts quickly.

Our aim is to tap the potential of the EU resources so that they can be used by other non-EU languages that are closely related to one or more official languages of the EU. Examples of such EU–non-EU language pairs include Swedish–Norwegian, Bulgarian–Macedonian[1], Romanian-Moldovan[2] and some other. After Croatia joins the EU, Serbian, Bosnian, and Montenegrin[3] will also be able to benefit from Croatian gradually turning resource-rich (all four languages have split from Serbo-Croatian after the breakup of Yugoslavia in the 90's and remain mutually intelligible). The newly-made EU-official (and thus not as resource-rich) Czech and Slovak languages are another possible pair of candidates. Spanish–Catalan, Irish-Gaelic Scottish, Standard German–Swiss German, and Italian–Maltese[4] are other good examples. As we will see below, even such resource-rich languages like Spanish and Portuguese can benefit from the proposed approach. Of course, many pairs of closely related languages that could make use of each other's bi-texts can also be found outside of Europe: one such example is Malay–Indonesian, with which we will be experimenting below. Other non-EU language pairs that could potentially benefit include Modern Standard Arabic–Dialectical Arabic (e.g., Egyptian, Levantine, Gulf, or Iraqi Arabic), Mandarin–Cantonese, Russian–Ukrainian, Turkish–Azerbaijani, Hindi–Urdu, and many other.

---

1. There is a heated linguistic debate about whether Macedonian represents a separate language or is a regional literary form of Bulgarian. Since there are no clear criteria for distinguishing a dialect from a language, linguists are divided on this issue. Politically, the Macedonian language is not recognized by Bulgaria (which refers to it as "*the official language of the Republic of Macedonia in accordance with its constitution*") and by Greece (mostly because of the dispute over the use of the name *Macedonia*).

2. As with Macedonian, there is a debate about the existence of the Moldovan language. While linguists generally agree that Moldovan is one of the dialects of Romanian, politically, the national language of Moldova can be called both *Moldovan* and *Romanian*.

3. There is a serious internal political division in Montenegro on whether the national language should be called *Montenegrin* or just *Serbian*.

4. Though, Maltese might benefit from Arabic more than from Italian.

Below we propose using bi-texts for resource-rich language pairs to build better SMT systems for resource-poor pairs by exploiting the similarity between a resource-poor language and a resource-rich one. More precisely, we build phrase-based SMT systems that translate from a resource-poor language $X_1$ into a resource-rich language $Y$ given a small bi-text for $X_1$-$Y$ and a much larger bi-text for $X_2$-$Y$, where $X_1$ and $X_2$ are closely related.

We are motivated by the observation that related languages tend to have (1) similar word order and syntax, and, more importantly, (2) overlapping vocabulary, e.g., *casa* ('house') is used in both Spanish and Portuguese; they also have (3) similar spelling. This vocabulary overlap means that the resource-rich auxiliary language can be used as a source of translation options for words that cannot be translated with the resources available for the resource-poor language. In actual text, the vocabulary overlap might extend from individual words to short phrases (especially if the resource-rich languages has been transliterated to look like the resource-poor one), which means that translations of whole phrases could potentially be reused between related languages. Moreover, the vocabulary overlap and the similarity in word order can be used to improve the word alignments for the resource-poor language by biasing the word alignment process with additional sentence pairs from the resource-rich language. We take advantage of all these opportunities: (1) we improve the word alignments for the resource-poor language, (2) we further augment it with additional translation options, and (3) we take care of potential spelling differences through appropriate transliteration.

We apply our approach to Indonesian→English using Malay and to Spanish→English using Portuguese and Italian (and pretending that Spanish is resource-poor), achieving sizable performance gains (up to 3.37 BLEU points) when using additional bi-texts for a related resource-rich language. We further show that our approach outperforms the best rivaling approaches, while using less additional data. Overall, we cut the amount of necessary "real" training data by a factor of 2–5.

Our approach is based on the phrase-based SMT model (Koehn, Och, & Marcu, 2003), which is the most commonly used state-of-the-art model today. However, the general ideas can easily be extended to other SMT models, e.g., hierarchical (Chiang, 2005), treelet (Quirk, Menezes, & Cherry, 2005), and syntactic (Galley, Hopkins, Knight, & Marcu, 2004).

The remainder of this article is organized as follows: Section 2 provides an overview of related work, Section 3 presents a motivating example in several languages, Section 4 introduces our proposed approach and discusses various alternatives, Section 5 describes the datasets we use, Section 6 explains how we transliterate Portuguese and Italian to look like Spanish automatically, Section 7 presents our experiments and discusses the results, Section 8 analyses the results in more detail, and, finally, Section 9 concludes and suggests possible directions for future work.

## 2. Related Work

Our general problem formulation is a special case of *domain adaptation*. Moreover, there are three basic concepts that are central to our work: (1) *cognates* between related languages, (2) machine translation between *closely related languages*, and (3) *pivoting* for statistical machine translation. We will review the previous work on these topics below, while also mentioning some other related work whenever appropriate.

## 2.1 Domain Adaptation

The *Domain adaptation* (or *transfer learning*) problem arises in situations where the training and the test data come from different distributions, thus violating the fundamental assumption of statistical learning theory. Our problem is an instance of the special case of domain adaptation, where in-domain data is scarce, but there is plenty of out-of-domain data. Many efficient techniques have been developed for domain adaptation in natural language processing; see the work of Daumé and Marcu (2006), Jiang and Zhai (2007a, 2007b), Chan and Ng (2005, 2006, 2007), and Dahlmeier and Ng (2010) for some examples.

Unfortunately, these techniques are not directly applicable to machine translation, which is much more complicated, and leaves a lot more space for variety in the proposed solutions. This is so despite the limited previous work on domain adaptation for SMT, which has focused almost exclusively on adapting European parliament debates to the news domain as part of the annual competition on machine translation evaluation at the WMT workshop. To mention just a few of the proposed approaches, Hildebrand, Eck, Vogel, and Waibel (2005) use information retrieval techniques to choose training samples that are similar to the test set as a way to adapt the translation model, while Ueffing, Haffari, and Sarkar (2007) adapt the translation model in a semi-supervised manner using monolingual data from the source language. Snover, Dorr, and Schwartz (2008) adapt both the translation and the language model, using comparable monolingual data in the target language. Nakov and Ng (2009b) adapt the translation model for phrase-based SMT by combining phrase tables using extra features indicating the source of each phrase; we will use this combination technique as part of our proposed approach below. Finally, Daumé and Jagarlamudi (2011) address the domain shift problem by mining appropriate translations for the unseen words.

## 2.2 Cognates

Cognates are defined as pairs of source-target words with similar spelling (and thus likely similar meaning), for example, *développement* in French *vs. development* in English. Many researchers have used likely cognates co-occurring in parallel sentences in the training bi-text to improve word alignments and ultimately build better SMT systems.

Al-Onaizan, Curin, Jahr, Knight, Lafferty, Melamed, Och, Purdy, Smith, and Yarowsky (1999) extracted such likely cognates for Czech-English, using one of the variations of the *longest common subsequence ratio* or LCSR (Melamed, 1995) described by Tiedemann (1999) as a similarity measure. They used these cognates to improve word alignments with IBM models 1–4 in three different ways: (1) by seeding the parameters of IBM model 1, (2) by constraining the word co-occurrences when training IBM models 1–4, and (3) by adding the cognate pairs to the bi-text as additional "sentence pairs". The last approach performed best and was later used by Kondrak, Marcu, and Knight (2003) who demonstrated improved SMT for nine European languages. It was further extended by Nakov, Nakov, and Paskaleva (2007), who combined LCSR and sentence-level co-occurrences in a bi-text with *competitive linking* (Melamed, 2000), language-specific weights, and Web $n$-gram frequencies.

Unlike these approaches, which extract cognates between the source and the target language, we use cognates between the source and some other related language that is different from the target. Moreover, we only implicitly rely on the existence of such cognates;

we do not try to extract them at all, and we leave them in their original sentence contexts.[5] Note that our approach is orthogonal to this kind of cognate extraction from the original training bi-text, and thus the two can be combined (which we will do in Section 7.7).

Another relevant line of research is on using cognates to adapt resources for one language to another one. For example, Hana, Feldman, Brew, and Amaral (2006) adapt Spanish resources to Brazilian Portuguese to train a part-of-speech tagger.

Cognates and cognate extraction techniques have been used in many other applications, e.g., for automatic translation lexicon induction. For example, Mann and Yarowsky (2001) induce translation lexicons between a resource-rich language (e.g., English) and a resource-poor language (e.g., Portuguese) using a resource-rich bridge language that is closely related to the latter (e.g., Spanish). They use pre-existing translation lexicons for the source-to-bridge mapping step (e.g., English-Spanish), and string distance measures for finding cognates for the bridge-to-target step (e.g., Spanish-Portuguese). This work was extended by Schafer and Yarowsky (2002), and later by Scherrer (2007), who relies on graphemic similarity for inducing bilingual lexicons between Swiss German and Standard German.

Koehn and Knight (2002) describe several techniques for inducing translation lexicons from monolingual corpora. Starting with unrelated German and English corpora, they look for (1) identical words, (2) cognates, (3) words with similar frequencies, (4) words with similar meanings, and (5) words with similar contexts. This is a bootstrapping process, where new translation pairs are added to the lexicon at each iteration.

More recent work on automatic lexicon induction includes that by Haghighi, Liang, Berg-Kirkpatrick, and Klein (2008), and Garera, Callison-Burch, and Yarowsky (2009).

Finally, there is a lot of research on string similarity that has been applied to cognate identification: Ristad and Yianilos (1998) and Mann and Yarowsky (2001) use the *minimum edit distance ratio* or MEDR with weights that are learned automatically using a stochastic transducer. Tiedemann (1999) and Mulloni and Pekar (2006) learn automatically the regular spelling changes between two related languages, which they incorporate in similarity measures based on LCSR and on MEDR, respectively. Kondrak (2005) proposes a formula for measuring string similarity based on LCSR with a correction that addresses its general preference for short words. Klementiev and Roth (2006) and Bergsma and Kondrak (2007) propose discriminative frameworks for measuring string similarity. Rappoport and Levent-Levi (2006) learn substring correspondences for cognates, using the string-level substitutions framework of Brill and Moore (2000). Finally, Inkpen, Frunza, and Kondrak (2005) compare several orthographic similarity measures for cognate extraction.

While cognates are typically extracted between related languages, there are words with similar spelling between unrelated languages as well, e.g., Arabic, Chinese, Japanese, and Korean proper names are *transliterated* to English, which uses a different alphabet. See the work of Oh, Choi, and Isahara (2006) for an overview and a comparison of different transliteration models, as well as the proceedings of the annual NEWS named entities workshop, which features shared tasks on transliteration mining and generation (Li & Kumaran, 2010). Transliteration can be modeled using character-based machine translation techniques (Matthews, 2007; Nakov & Ng, 2009a; Tiedemann & Nabende, 2009), which are related to the character-based SMT model of Vilar, Peter, and Ney (2007), and Tiedemann (2009).

---

5. However, in some of our experiments, we extract cognates for training a transliteration system from the resource-rich source language $X_2$ into the resource-poor one $X_1$.

## 2.3 Machine Translation between Closely Related Languages

Yet another relevant line of research is on machine translation between closely related languages, which is arguably simpler than general SMT, and thus can be handled using word-for-word translation and manual language-specific rules that take care of the necessary morphological and syntactic transformations. This has been tried for a number of language pairs including Czech–Slovak (Hajič, Hric, & Kuboň, 2000), Turkish–Crimean Tatar (Altintas & Cicekli, 2002), and Irish–Scottish Gaelic (Scannell, 2006), among others. More recently, the Apertium open-source machine translation platform at `http://www.apertium.org/` has been developed, which uses bilingual dictionaries and manual rules to translate between a number of related languages, including Spanish–Catalan, Spanish–Galician, Occitan–Catalan, and Macedonian-Bulgarian. In contrast, we have a language-independent, statistical approach, and a different objective: translate into a third language $X$.

A special case of this same line of research is the translation between dialects of the same language, e.g., between Cantonese and Mandarin (Zhang, 1998), or between a dialect of a language and a standard version of that language, e.g., between some Arabic dialect (e.g., Egyptian) and Modern Standard Arabic (Bakr, Shaalan, & Ziedan, 2008; Sawaf, 2010; Salloum & Habash, 2011). Here again, manual rules and/or language-specific tools are typically used. In the case of Arabic dialects, a further complication arises by the informal status of the dialects, which are not standardized and not used in formal contexts but rather only in informal online communities[6] such as social networks, chats, Twitter and SMS messages. This causes further mismatch in domain and genre.

Thus, translating from Arabic dialects to Modern Standard Arabic requires, among other things, normalizing informal text to a formal form. In fact, this is a more general problem, which arises with informal sources like SMS messages and Tweets for any language (Han & Baldwin, 2011). Here the main focus is on coping with spelling errors, abbreviations, and slang, which are typically addressed using string edit distance, while also taking pronunciation into account. This is different from our task, where we try to reuse good, formal text from one language to help improve SMT for another language.

A closely related relevant line of research is on language adaptation and normalization, when done specifically for improving SMT into another language. For example, Marujo, Grazina, Luís, Ling, Coheur, and Trancoso (2011) described a rule-based system for adapting Brazilian Portuguese (BP) to European Portuguese (EP), which they used to adapt BP–English bi-texts to EP–English. Unlike this work, which heavily relied on language-specific rules, our approach is statistical and largely language-independent; more importantly, we have a different objective: translate into a third language $X$.

## 2.4 Pivoting

Another relevant line of research is improving SMT using additional languages as pivots.

Callison-Burch, Koehn, and Osborne (2006) improved phrase-based SMT from Spanish and French to English using source-language phrase-level paraphrases extracted using the pivoting technique of Bannard and Callison-Burch (2005) and eight additional languages from the *Europarl corpus* (Koehn, 2005).

---

6. The Egyptian Wikipedia is one notable exception.

For example, using German as a pivot, they extracted English paraphrases from a parallel English-German bi-text by looking for English phrases that were aligned to the same German phrase: e.g., if *under control* and *in check* were aligned to *unter controlle*, they were hypothesized to be paraphrases with some probability. Such Spanish/French paraphrases were added as additional entries in the phrase table of an Spanish→English/French→English phrase-based SMT system and paired with the English translation of the original Spanish/French phrase. The system was then tuned with minimum error rate training (MERT) (Och, 2003), adding an extra feature penalizing low-probability paraphrases; this yielded huge increase in coverage (from 48% to 90% of the test word types when 10K training sentence pairs were used), and up to 1.8 BLEU points of absolute improvement.

Unlike this kind of pivoting, which can only improve source-language lexical coverage, we augment both the source- and the target-language sides. Second, while pivoting ignores context when extracting paraphrases, we do take it into account. Third, by using as an additional language one that is related to the source, we are able to get increase in BLEU that is comparable and even better than what pivoting achieves with eight pivot languages. On the negative side, our approach is limited in that it requires that the auxiliary language $X_2$ be related to source language $X_1$, while the pivoting language $Z$ does not have to be related to $X_1$ nor to the target language $Y$. However, we only need one additional parallel corpus (for $X_2$-$Y$), while pivoting needs two: one for $X_1$-$Z$ and one for $Z$-$Y$. Finally, note that our approach is orthogonal to pivoting, and thus the two can be combined (which we will do in Section 7.8).

We should note that pivoting is a more general technique, which has been widely used in statistical machine translation, e.g., for triangulation, where one wants to build a French-German machine translation system from a French-English and an English-German bi-text, without an access to a French-German bi-text. In that case, pivoting can be done at the sentence-level, e.g., by cascading translation systems, first translating from French to English, and then translating from English to German (de Gispert & Mario, 2006; Utiyama & Isahara, 2007) or at the phrase-level, e.g., using the phrase table composition, which can be done off-line (Cohn & Lapata, 2007; Wu & Wang, 2007), or it can be integrated in the decoder (Bertoldi, Barbaiani, Federico, & Cattoni, 2008). It has been also shown that pivoting can outperform direct translation, e.g., translating from Arabic to Chinese could work better using English as a pivot than if done directly (Habash & Hu, 2009). Moreover, it has been argued that English might not always be the optimal choice of a pivot language (Paul, Yamamoto, Sumita, & Nakamura, 2009). Finally, pivoting techniques have been also used at the word-level, e.g., for translation lexicon induction between Japanese and German using English (Tanaka, Murakami, & Ishida, 2009), or for improving word alignments (Filali & Bilmes, 2005; Kumar, Och, & Macherey, 2007). Pivot languages have also been used for lexical adaptation (Crego, Max, & Yvon, 2010).

Overall, all these more general pivoting techniques aim to build a machine translation system for a new (resource-poor) language pair $X$-$Y$, assuming the existence of bi-texts $X$-$Z$ and $Z$-$Y$ for some auxiliary pivoting language $Z$, e.g., they would be useful for translating between Malay and Indonesian, by pivoting over English. In contrast, we are interested in building a better system for translating not from $X$ to $Y$ but from $X$ to $Z$, e.g., from Indonesian to English. We further assume that the bi-text for $X$-$Z$ is small, while the one for $Z$-$Y$ is large, and we require that $X$ and $Y$ be closely related languages.

Another related line of research is on statistical multi-source translation, which focuses on translating a text given in multiple source languages into a single target language (Och & Ney, 2001; Schroeder, Cohn, & Koehn, 2009). This situation arises for a small number of resource-rich languages in the context of the United Nations or the European Union, but it could hardly be expected for resource-poor languages.

## 3. Motivating Example

Consider Article 1 of the *Universal Declaration of Human Rights*:

> *All human beings are born free and equal in dignity and rights. They are endowed with reason and conscience and should act towards one another in a spirit of brotherhood.*

and let us see how it is translated in the closely related Malay and Indonesian and the more dissimilar Spanish and Portuguese.

### 3.1 Malay and Indonesian

Malay (aka *Bahasa Malaysia*) and Indonesian (aka *Bahasa Indonesia*) are closely related Astronesian languages, with about 180 million speakers combined. Malay is official in Malaysia, Singapore and Brunei, and Indonesian is the national language of Indonesia. The two languages are mutually intelligible to a great extent, but they differ in orthography/pronunciation and vocabulary.

Malay and Indonesian use a unified spelling system based on the Latin alphabet, but they exhibit occasional differences in orthography due to diverging pronunciation, e.g., *kerana vs. karena* ('because') and *Inggeris vs. Inggris* ('English') in Malay and Indonesian, respectively. More rarely, the differences are historical, e.g., *wang vs. uang* ('money').

The two languages differ more substantially in vocabulary, mostly because of loan words, where Malay typically follows the English pronunciation, while Indonesian tends to follow Dutch, e.g., *televisyen vs. televisi, Julai vs. Juli*, and *Jordan vs. Yordania*. For words of Latin origin that end on *-y* in English, Malay uses *-i*, while Indonesian uses *-as*, e.g., *universiti vs. universitas, kualiti vs. kualitas*.

While there are many cognates between the two languages, there are also some *false friends*, which are words identically spelled but with different meanings in the two languages. For example, *polisi* means *policy* in Malay but *police* in Indonesian. There are also many partial cognates, e.g., *nanti* means both *will* (future tense marker) and *later* in Malay but only *later* in Indonesian. As a result, fluent Malay and fluent Indonesian can differ substantially. Consider, for example, the Malay and the Indonesian versions of Article 1 of the *Universal Declaration of Human Rights* (from the official website of the United Nations):

- **Malay:** ___Semua___ manusia ___dilahirkan___ bebas ___dan___ samarata dari segi kemuliaan ___dan hak-hak___. ___Mereka___ mempunyai pemikiran ___dan___ perasaan ___hati___ ___dan___ hendaklah bertindak di antara ___satu sama lain___ dengan ___semangat persaudaraan___.

- **Indonesian:** ___Semua___ orang ___dilahirkan___ merdeka ___dan___ mempunyai martabat ___dan hak-hak___ yang sama. ___Mereka___ dikaruniai akal ___dan___ ___hati___ nurani ___dan___ hendaknya bergaul ___satu sama lain___ dalam ___semangat persaudaraan___.

Semantically, the overlap is substantial, and a native speaker of Indonesian can understand most of what the Malay version says, but would find parts of it not quite fluent.

In the above example, there is only 50% overlap at the individual word level (overlapping words are underlined). In fact, the actual vocabulary overlap is much higher, e.g., there is only one word in the Malay text that does not exist in Indonesian: *samarata*. Other differences are due to the use of different morphological forms, e.g., *hendaklah vs. hendaknya* ('conscience'), both derivational variants of *hendak* ('want').

Of course, word choice in translation is often a matter of taste, and thus not all differences above are necessarily required. To test this, we asked a native speaker of Indonesian to adapt the Malay version to Indonesian while preserving as many words as possible. This yielded the following, arguably somewhat less fluent, Indonesian version, which only has six words that are not in the Malay version:

- **Indonesian (closer to Malay):** <u>Semua manusia dilahirkan bebas dan</u> mempunyai martabat <u>dan hak-hak</u> yang sama. <u>Mereka mempunyai pemikiran dan perasaan dan hendaklah</u> bergaul <u>satu sama lain</u> dalam <u>semangat persaudaraan</u>.

Note the increase in the average length of the matching phrases for this adapted version.

## 3.2 Spanish and Portuguese

Spanish and Portuguese also exhibit a noticeable degree of mutual intelligibility, but differ in pronunciation, spelling, and vocabulary. Unlike Malay and Indonesian, however, they also differ syntactically and exhibit a high level of spelling differences; this can be seen from the translation of Article 1 of the *Universal Declaration of Human Rights*:

- **Spanish:** ***Todos*** los ***seres humanos*** nacen libres ***e*** iguales en dignidad y derechos y, ***dotados*** como están de razón y conciencia, deben comportarse fraternalmente los unos con los otros.

- **Portuguese:** ***Todos*** os ***seres humanos*** nascem livres ***e*** iguais em dignidade e em direitos. ***Dotados*** de razão e de consciência, devem agir uns para com os outros em espírito de fraternidade.

We can see that the exact word-level overlap between the Spanish and the Portuguese is quite low: about 17% only. Still, we can see some overlap at the level of short phrases, not just at the word level.

Spanish and Portuguese share about 90% of their vocabulary and thus the observed level of overlap may appear surprisingly low. The reason is that many cognates between the two languages exhibit minor spelling variations. These variations can stem from different rules of orthography, e.g., *se<u>nh</u>or vs. se<u>ñ</u>or* in Portuguese and Spanish, but they can also be due to genuine phonological differences. For example, the Portuguese suffix *-ção* corresponds to the Spanish suffix *-ción*, e.g., *evolu<u>ção</u> vs. evolu<u>ción</u>*. Similar systematic differences exist for verb endings like *-ou vs. -ó* (for $3^{rd}$ person singular, simple past tense), e.g., *visit<u>ou</u> vs. visit<u>ó</u>*, or *-ei vs. -é* (for 1st person singular, simple past tense), e.g., *visit<u>ei</u> vs. visit<u>é</u>*. There are also occasional differences that apply to a particular word only, e.g., *di<u>z</u>er vs. de<u>c</u>ir*, *Mário vs. Mario*, and *Maria vs. María*.

Going back to our example, if we ignore the spelling variations between the cognates in the two languages, the overlap jumps significantly:

- **Portuguese (cognates transliterated to Spanish):**
  ***Todos los seres humanos nacen libres e iguales en dignidad y en derechos***. ***Dotados*** *de* ***razón y*** *de* ***conciencia, deben*** *agir* ***unos*** *para* ***con los otros*** *en espírito de fraternidad.*

All words in the above sentence are Spanish, and most of the differences from the official Spanish version above are due to different word choice by the translator; in fact, the sentence can become fluent Spanish if *agir unos par* is changed to *comportarse los unos con.*

## 4. Method

The above examples suggest that it may be feasible to use bi-texts for one language to improve SMT for some related language, possibly after suitable transliteration of the cognates in the additional language to match the target spelling.

Thus, below we describe two general strategies for improving phrase-based SMT from some resource-poor language $X_1$ into a target language $Y$, using a bi-text $X_2$-$Y$ for a related resource-rich language $X_2$: (a) bi-text concatenation, with possible repetitions of the original bi-text for balance, and (b) phrase table combination, where each bi-text is used to build a separate phrase table, and then the two phrase tables are combined. We discuss the advantages and disadvantages of these general strategies, and we propose a hybrid approach that combines their strengths while trying to avoid their limitations.

### 4.1 Concatenating Bi-texts

We can simply concatenate the bi-texts for $X_1$-$Y$ and $X_2$-$Y$ into one large bi-text and use it to train an SMT system. This offers several potential benefits.

First, it can yield improved word alignments for the sentences that came from the $X_1$-$Y$ bi-text, e.g., since the additional sentences can provide new contexts for the rare words in that bi-text, thus potentially improving their alignments, which in turn could yield better phrase pairs. Rare words are known to serve as "garbage collectors" (Brown, Della Pietra, Della Pietra, Goldsmith, Hajič, Mercer, & Mohanty, 1993) in the IBM word alignment models. Namely, a rare source word tends to align to many target language words rather than allowing them to stay unaligned or to align to other source words. The problem is not limited to IBM word alignment models (Brown, Della Pietra, Della Pietra, & Mercer, 1993); it also exists for the HMM model of Vogel, Ney, and Tillmann (1996). See Graca, Ganchev, and Taskar (2010) for a detailed discussion and examples of the "garbage collector effect".

Moreover, concatenation can provide new source-language side translation options, thus increasing lexical coverage and reducing the number of unknown words; it can also provide new useful non-compositional phrases on the source-language side, thus yielding more fluent translation output. It also offers new target-language side phrases for known source phrases, which could improve fluency by providing more translation options for the language model to choose from. Finally, inappropriate phrases including words from $X_2$ that do not exist in $X_1$ will not match the test-time input, while inappropriate new target-language translations still have the chance to be filtered out by the language model.

However, simple concatenation can be problematic. First, when concatenating the small bi-text for $X_1$-$Y$ with the much larger one for $X_2$-$Y$, the latter will dominate during word alignment and phrase extraction, thus hugely influencing both lexical and phrase translation probabilities, which can yield poor performance. This can be counter-acted by repeating the small bi-text several times so that the large one does not dominate. Second, since the bi-texts are merged mechanically, there is no way to distinguish between phrases extracted from the bi-text for $X_1$-$Y$ from those coming from the bi-text for $X_2$-$Y$. The former are for the target language pair and thus probably should be preferred, while using the latter should be avoided since they might contain inappropriate translations for some words from $X_1$. For example, a phrase pair from the Indonesian-English bi-text could (correctly) translate *polisi* as *police*, while one from the Malay-English bi-text could (correctly for Malay, but inappropriately for Indonesian) translate it as *policy*. This is because the Malay word *polisi* and the Indonesian word *polisi* are false friends.

We experiment with combining the original and the additional training bi-text in the following three ways:

- **cat×1**: We simply concatenate the original and the additional training bi-text to form a new training bi-text, which we use to train a phrase-based SMT system.

- **cat×k**: We concatenate $k$ copies of the original and one copy of the additional training bi-text to form a new training bi-text. The value of $k$ is selected so that the original bi-text approximately matches the size of the additional bi-text.

- **cat×k:align**: We concatenate $k$ copies of the original and one copy of the additional training bi-text to form a new training bi-text. We generate word alignments for this concatenated bi-text. Then we throw away all sentence pairs and their alignments, except for one copy of the original bi-text. Thus, effectively we induce word alignments for the original bi-text only, while using the concatenated bi-text to estimate the statistics about them. We then use these alignments to build a phrase table for the original bi-text.

The first and the second method represent simple and balanced bi-text concatenation, respectively. The third method is a version of the second one, where the additional bi-text is only used to improve the word alignments for the original bi-text, but is not used for phrase extraction. Thus, it isolates the effect of improved word alignments from the effect of improved vocabulary coverage that the additional training bi-text can provide. **cat×1** and **cat×k:align** will be the basic building blocks of our more sophisticated approach below.

## 4.2 Combining Phrase Tables

An alternative way of making use of the additional training bi-text for the resource-rich language pair $X_2$-$Y$ in order to train an improved phrase-based SMT system for $X_1 \rightarrow Y$ is to build separate phrase tables from $X_1$-$Y$ and $X_2$-$Y$, which can then be (a) used together, e.g., as alternative decoding paths, (b) merged, e.g., using one or more extra features to indicate the bi-text each phrase pair came from, or (c) interpolated, e.g., using simple linear interpolation.

Building two separate phrase tables offers several advantages. First, the preferable phrase pairs extracted from the bi-text for $X_1$-$Y$ are clearly distinguished from (or given a higher weight in the linear interpolation compared to) the potentially riskier ones from the $X_2$-$Y$ bi-text. Second, the lexical and the phrase translation probabilities are combined in a principled manner. Third, using the $X_2$-$Y$ bi-text, which is much larger than that for $X_1$-$Y$ is not problematic any more: it will not dominate as was the case with simple concatenation above. Finally, as with bi-text merging, there are many additional source- and target-language phrases, which offer new translation options. On the negative side, the opportunity is lost to improve word alignments for the sentences in the $X_1$-$Y$ bi-text.

We experiment with the following three phrase table combination strategies:

- **Two-tables**: We build two separate phrase tables, one for each of the two bi-texts, and we use them as alternative decoding paths (Birch, Osborne, & Koehn, 2007).

- **Interpolation**: We build two phrase tables, $T_{orig}$ and $T_{extra}$, for the original and for the additional bi-text, respectively, and we use linear interpolation to combine the corresponding conditional probabilities: $\Pr(e|s) = \alpha \Pr_{orig}(e|s) + (1 - \alpha) \Pr_{extra}(e|s)$. We optimize the value of $\alpha$ on the development dataset, *i.e.*, we run MERT for merged phrase tables generated using different values of $\alpha$, and we choose the value that gives rise to the phrase table that achieves the highest tuning BLEU score. In order to reduce the search space, we only try five values for $\alpha$ (.5, .6, .7, .8 and .9), *i.e.*, we reduce the tuning to this discrete set, and we use the same $\alpha$ for all four conditional probabilities in the phrase table.

- **Merge**: We build two separate phrase tables, $T_{orig}$ and $T_{extra}$, for the original and for the additional training bi-text, respectively. We then concatenate them, giving priority to $T_{orig}$ as follows: We keep all source-target phrase pairs from $T_{orig}$, adding to them those source-target phrase pairs from $T_{extra}$ that were not present in $T_{orig}$. For each source-target phrase pair added, we retain its associated conditional probabilities (forward/reverse phrase translation probability, and forward/reverse lexicalized phrase translation probability) and the phrase penalty.[7] We further add up to three additional features to each entry in the new table: $F_1$, $F_2$, and $F_3$. The value of $F_1$ is 1 if the source-target phrase pair originated from $T_{orig}$, and 0.5 otherwise. Similarly, $F_2$=1 if the source-target phrase pair came from $T_{extra}$, and $F_2$=0.5 otherwise. The value of $F_3$ is 1 if the source-target phrase pair was in both $T_{orig}$ and $T_{extra}$, and 0.5 otherwise. Thus, there are three possible feature value combinations: (1;0.5;0.5), (0.5;1;0.5) and (1;1;1); the last one is used for a phrase pair that was in both $T_{orig}$ and $T_{extra}$. We experiment with using (1) $F_1$ only, (2) $F_1$ and $F_2$, and (3) $F_1$, $F_2$, and $F_3$. We set the weights for all phrase table features, including the standard five and the additional three, using MERT. We further optimize the number of additional features (one, two, or three) on the development set, *i.e.*, we run MERT for phrase tables with one, two, and three extra features and we choose the phrase table that has achieved the highest BLEU score on tuning, as suggested in the work of Nakov (2008).

---

7. In theory, we should also re-normalize the probabilities since they may not sum to one. In practice, this is not that important since the log-linear phrase-based SMT model does not require that the features be probabilities at all, e.g., $F_1$, $F_2$, $F_3$, and the phrase penalty are not probabilities.

### 4.3 Proposed Approach

Taking into account the potential advantages and disadvantages of the above two general strategies, we propose an approach that tries to get the best from each of them, namely: ($i$) improved word alignments for $X_1$-$Y$, by biasing the word alignment process with additional sentence pairs from $X_2$-$Y$, and ($ii$) increased lexical coverage, by using additional phrase pairs that the $X_2$-$Y$ bi-text can provide. This is achieved by using **Merge** to combine the phrase tables for **cat**×$k$:**align** and **cat**×**1**. The process can be described in more detail as follows:

1. Build a balanced bi-text $B_{rep}$, which consists of the $X_1$-$Y$ bi-text repeated $k$ times followed by one copy of the $X_2$-$Y$ bi-text. Generate word alignments for $B_{rep}$, then truncate them, only keeping word alignments for one copy of the $X_1$-$Y$ bi-text. Use these word alignments to extract phrases, and build a phrase table $T_{rep\_trunc}$.

2. Build a bi-text $B_{cat}$ that is a simple concatenation of the bi-texts for $X_1$-$Y$ and $X_2$-$Y$. Generate word alignments for $B_{cat}$, extract phrases, and build a phrase table $T_{cat}$.

3. Generate a merged phrase table by combining $T_{rep\_trunc}$ and $T_{cat}$. The merging gives priority to $T_{rep\_trunc}$ and uses extra features indicating the origin of each entry in the combined phrase table.

## 5. Datasets

We experiment with the following bi-texts and monolingual English data:

- **Indonesian-English** (*in-en*)**:**

  - train: 28,383 sentence pairs (0.8M, 0.9M words);
  - dev: 2,000 sentence pairs (56.6K, 63.3K words);
  - test: 2,000 sentence pairs (58.2K, 65.0K words);
  - monolingual English $en_{in}$: 5.1M words.

- **Malay-English** (*ml-en*)**:**

  - train: 190,503 sentence pairs (5.4M, 5.8M words);
  - dev: 2,000 sentence pairs (59.7K, 64.5K words);
  - test: 2,000 sentence pairs (57.9K, 62.4K words);
  - monolingual English $en_{ml}$: 27.9M words.

- **Spanish-English** (*es-en*)**:**

  - train: 1,240,518 sentence pairs (35.7M, 34.6M words);
  - dev: 2,000 sentence pairs (58.9K, 58.1K words);
  - test: 2,000 sentence pairs (56.2K, 55.5K words);
  - monolingual English $en_{es:pt}$: 45.3M words (the same as for *pt-en* and *it-en*).

- **Portuguese-English** (*pt-en*):

    - train: 1,230,038 sentence pairs (35.9M, 34.6M words).
    - dev: 2,000 sentence pairs (59.3K, 58.5K words);
    - test: 2,000 sentence pairs (56.5K, 55.7K words);
    - monolingual English $en_{es:pt}$: 45.3M words (the same as for *es-en* and *it-en*).

- **Italian-English** (*it-en*):

    - train: 1,565,885 sentence pairs (43.5M, 44.1M words);
    - dev: 2,000 sentence pairs (56.8K, 57.7K words);
    - test: 2,000 sentence pairs (57.4K, 60.3K words);
    - monolingual English $en_{es:it}$: 45.3M words (the same as for *es-en* and *pt-en*).

The lengths of the sentences in all bi-texts above are limited to 100 tokens. For each of the language pairs, we have a development and a testing bi-text, each with 2,000 parallel sentence pairs. We made sure the development and the testing bi-texts shared no sentences with the training bi-texts; we further excluded from the monolingual English data all sentences from the English sides of the development and the testing bi-texts.

The training bi-text datasets for *es-en*, *pt-en*, and *it-en* were built from v.3 of the *Europarl* corpus, excluding the Q4/2000 portion of the data (2000-10 to 2000-12), out of which we created our testing and development datasets.

We built the *in-en* bi-texts from comparable texts that we downloaded from the Web. We translated the Indonesian texts to English using *Google Translate*, and we matched[8] them against the English texts using a cosine similarity measure and heuristic constraints based on document length in words and in sentences, overlap of numbers, words in uppercase, and words in the title. Next, we extracted pairs of sentences from the matched document pairs using *competitive linking* (Melamed, 2000), and we retained the ones whose similarity was above a pre-specified threshold. The *ml-en* bi-text was built similarly.

For all pairs of languages, the monolingual English text for training the language model consists of the English side of the corresponding bi-text plus some additional English text from the same source.

Note that the monolingual data for training an English language model is the same for Spanish, Portuguese, and Italian since the *es-en*, *pt-en*, and *it-en* are from the same origin: in fact, with very few exceptions, the sentences in these bi-texts can be aligned over English to make a *es-en-pt-it* four-text, since they are all translations (from English and other languages) of the same original parliamentary debates. Thus, the English side of *es-en*, *pt-en*, and *it-en*, and of the unaligned English sentences have the same distribution.

This is not the case, however, for Malay and Indonesian, which come from different sources and are on different topics – they discuss issues in Malaysia and Indonesia, respectively. In particular, they differ a lot in the use of named entities: names of persons, locations, and organizations that they talk about. This is why we have separate monolingual texts to train English language models for *ml-en* and *in-en*; as we will see below, they do indeed yield different performance for SMT.

---

8. Note that the automatic translations were used for matching only; the final bi-text contained no automatic translations.

## 6. Transliteration

As we mentioned above, our approach relies on the existence of a large number of *cognates* between related languages. While linguists define cognates as words derived from a common root[9] (Bickford & Tuggy, 2002), *computational* linguists typically ignore origin, defining them as words in different languages that are mutual translations and have a similar orthography (Melamed, 1999; Mann & Yarowsky, 2001; Bergsma & Kondrak, 2007). Here we adopt the latter definition.

As we have seen in Section 3, transliteration can be very helpful for languages like Spanish and Portuguese, which have many regular spelling differences. Thus, we build a system for automatic transliteration from Portuguese to Spanish, which we train on a list of automatically extracted pairs of likely cognates. We apply this system on the Portuguese side of the *pt-en* training bi-text.

Classic approaches to automatic cognate extraction look for non-stopwords with similar spelling that appear in parallel sentences in a bi-text (Kondrak et al., 2003). In our case, however, we need to extract cognates between Spanish and Portuguese given *pt-en* and *es-en* bi-texts only, *i.e.*, without having a *pt-es* bi-text. Although it is easy to construct a *pt-es* bi-text from the Europarl corpus, we chose not to do so since, in general, synthesizing a bi-text for $X_1$-$X_2$ would be impossible: e.g., it cannot be done for *ml-in* given our training datasets for *in-en* and *ml-en* since their English sides have no sentences in common.

Thus, we extracted the list of likely cognates between Portuguese and Spanish from the training *pt-en* and *es-en* bi-texts using English as a pivot as follows: We started with IBM model 4 word alignments, from which we extracted four conditional lexical translation probabilities: $\Pr(p_j|e_i)$ and $\Pr(e_i|p_j)$ for Portuguese-English, and $\Pr(s_k|e_i)$ and $\Pr(e_i|s_k)$ for Spanish-English, where $p_j$, $e_i$, and $s_k$ stand for a Portuguese, an English and a Spanish word, respectively. Following Wu and Wang (2007), we then induced conditional lexical translation probabilities $\Pr(p_j|s_k)$ and $\Pr(s_k|p_j)$ for Portuguese-Spanish as follows:

$$\Pr(p_j|s_k) = \sum_i \Pr(p_j|e_i, s_k) \Pr(e_i|s_k)$$

Assuming $p_j$ is conditionally independent of $s_k$ given $e_i$, we can simplify this:

$$\Pr(p_j|s_k) = \sum_i \Pr(p_j|e_i) \Pr(e_i|s_k)$$

Similarly, for $\Pr(s_k|p_j)$, we obtain

$$\Pr(s_k|p_j) = \sum_i \Pr(s_k|e_i) \Pr(e_i|p_j)$$

We excluded all stopwords, words of length less than three, and those containing digits. We further calculated $\text{Prod}(p_j, s_k) = \Pr(p_j|s_k) \Pr(s_k|p_j)$, and we excluded all Portuguese-Spanish word pairs $(p_j, s_k)$ for which $\text{Prod}(p_j, s_k) < 0.01$. The value of 0.01 has been previously suggested for filtering phrase pairs obtained using pivoting (Callison-Burch, 2008, 2012; Denkowski & Lavie, 2010; Denkowski, 2012). From the remaining pairs, we extracted likely cognates based on $\text{Prod}(p_j, s_k)$ and on the orthographic similarity between $p_j$ and $s_k$.

Following Melamed (1995), we measured the orthographic similarity using the *longest common subsequence ratio* (LCSR), defined as follows:

---

9. E.g., Latin *tu*, Old English *thou*, Greek *sú*, and German *du* are all cognates meaning '$2^{nd}$ person singular'.

$$\text{LCSR}(s_1, s_2) = \frac{|\text{LCS}(s_1, s_2)|}{\max(|s_1|, |s_2|)}$$

where $\text{LCS}(s_1, s_2)$ is the *longest common subsequence* of $s_1$ and $s_2$, and $|s|$ is the length of $s$.

We retained as likely cognates all pairs for which LCSR was 0.58 or higher; this value was found by Kondrak et al. (2003) to be optimal for a number of language pairs in the *Europarl* corpus.

Finally, we performed *competitive linking* (Melamed, 2000), assuming that each Portuguese wordform had at most one Spanish best cognate match. Thus, using the values of $\text{Prod}(p_j, s_k)$, we induced a fully-connected weighted bipartite graph. Then, we performed a greedy approximation to the maximum weighted bipartite matching in that graph, *i.e.*, competitive linking, as follows: First, we accepted as cognates the cross-lingual pair $(p_j, s_k)$ with the highest $\text{Prod}(p_j, s_k)$ in the graph, and we discarded the words $p_j$ and $s_k$ from further consideration. Then, we accepted the next highest-scored pair, and we discarded the involved wordforms and so forth. The process was repeated until there were no matchable word pairs left.

Note that our cognate extraction algorithm has three components: (1) orthographic, based on LCSR, (2) semantic, based on pivoting over English, and (3) competitive linking.

The semantic component is very important and makes the extraction of "false friends" very unlikely. Consider for example the Spanish-Portuguese word pairs *largo – largo* and *largo – longo*. The latter is a pair of true cognates, but the former is a pair of "false friends" since *largo* means *long* in Spanish but *wide* in Portuguese. The word *largo* appears 8,489 times in the *es-en* bi-text and 432 times in the *pt-en* bi-text. However, having different meanings, they do not get aligned to the same English word with high probability, which results in very low scores for the conditional probabilities: $\Pr(p_j|s_k) = 0.000464$ and $\Pr(s_k|p_j) = 0.009148$; thus, $\text{Prod}(p_j, s_k) = 0.000004$, which is below the 0.01 threshold. As a result, the "false friend" pair *largo – largo* does not get extracted. In contrast, the true cognate pair *largo – longo* does get extracted because the corresponding conditional probabilities for it are 0.151354 and 0.122656, respectively, and their product is 0.018564, which is above 0.01 (moreover, LCSR = 0.6, which is above the 0.58 threshold).

The competitive linking component helps prevent issues related to word inflection that cannot be handled using pivoting alone. For example, the word for *green* in both Spanish and Portuguese has two forms: *verde* for singular, and *verdes* for plural. Without competitive linking, we would extract not only *verde – verde* ($\text{Prod}(p_j, s_k) = 0.353662$) and *verdes – verdes* ($\text{Prod}(p_j, s_k) = 0.337979$), but also the incorrect word pairs *verde – verdes* ($\text{Prod}(p_j, s_k) = 0.109792$) and *verdes – verde* ($\text{Prod}(p_j, s_k) = 0.106088$). Competitive linking, however, prevents this by asserting that no Portuguese and no Spanish word can have more than one true cognate, which effectively eliminates the wrong pairs.

Thus, taken together, the semantic component and competitive linking make the extraction of "false friends" very unlikely. Still, occasionally, we do get some wrong alignments such as *intrusa – intrusas*, where a singular form is matched with a plural form, which occurs mostly in the case of rare words like *intrusa* ('intruder', feminine) whose alignments tend to be unreliable, and for which very few inflected forms are available to competitive linking to choose from.

Note that the described transliteration system is focusing more on precision and less on recall. This is because the extracted likely cognate pairs are going to be used to train an SMT-based transliteration system. This system will have a translation component, which should be able to generate many options, and a target language model component, which would help filter those options. The translation component should tend to generate good options, and thus it needs to be trained primarily on instances of *systematic, regular* differences, such as *evolução – evolución*, from which the suffix change *-ção – -ción* can be learned. Occasional differences such as *dizer – decir* cannot be generalized and thus are less useful (they are also less frequent, and thus missing some of them is arguably not so important), but they can be simply memorized by the model as whole words and still used.

We should also note that our focus on precision of cognate pair extraction does not mean that we are going to extract primarily cognate pairs with very few spelling differences. As we explained above, spelling is just one component of our cognate pair extraction approach; there are also a semantic and a competitive linking component, which could eliminate many candidates with close spelling and prefer others with more dissimilarities (recall the correct choice of *largo – longo* over the wrong *largo – largo*).

Note that the generality of our transliteration approach is not necessarily compromised by the fact that LCSR requires that the languages use the same writing system. For example, Cyrillic-written Serbian and Roman-written Croatian can still be compared using LCSR, after an initial letter-by-letter mapping between the Cyrillic and the Roman alphabets, which is generally straightforward. Of course, even when using the same alphabet, languages can have different orthographical conventions, which might make them look more divergent than what the actual phonetics would suggest, e.g., compare *qui/chi, gui/ghi, glio/llo* in Spanish and Italian. Even though LCSR between the Italian-Spanish cognates *chi* and *qui* is lower than our threshold of 0.58, the correspondence between them as strings can still be learned from longer cognates, e.g., *macchina* and *máquina*. This would then allow the transliteration system to convert *chi* into *qui* as a word.

Going back to the actual experiments, as a result of the cognate extraction procedure, we ended up with 28,725 Portuguese-Spanish cognate pairs, 9,201 (or 32.03%) of which had spelling differences. For each pair in the list of cognate pairs, we added spaces between any two adjacent letters for both wordforms, and we further appended the start and the end characters ^ and $. For example, the cognate pair *evolução – evolución* became

$$\text{^ e v o l u ç ã o \$ — ^ e v o l u c i ó n \$}$$

We randomly split the resulting list into a training (26,725 pairs) and a development dataset (2,000 pairs), and we trained and tuned a character-level phrase-based monotone SMT system similar to Finch and Sumita (2008) to transliterate a Portuguese wordform into a Spanish wordform. We used a Spanish language model trained on 14M word tokens (obtained from the above-mentioned 45.3M-token monolingual English corpus after excluding punctuation, stopwords, words of length less than three, and those containing digits): one per line and character-separated with added start and end characters as in the above example. We set both the maximum phrase length and the language model order to ten; we found these values by tuning on the development dataset. We tuned the system using MERT, and we saved the feature weights. The tuning BLEU was 95.22%, while the baseline BLEU, for leaving the Portuguese words intact, was 87.63%.

Finally, we merged the training and the tuning datasets and we retrained. We used the resulting system with the saved feature weights to transliterate the Portuguese side of the training *pt-en* bi-text, which yielded a new *$pt_{es}$-en* training bi-text.

We repeated the same procedure for Italian-English. We extracted 25,107 Italian-Spanish cognate pairs, 14,651 (or 58.35%) of which had spelling differences. Then, we split the list into a training (23,107 pairs) and a development dataset (2,000 pairs), and trained a character-level phrase-based monotone SMT system as we did for Spanish-English; the tuning BLEU was 94.92%. We used the resulting system to transliterate the Italian side of the training *it-en* bi-text, thus obtaining a new *$it_{es}$-en* training bi-text.

We also applied transliteration to Malay into Indonesian, even though we knew that the spelling differences between these two languages were rare. We extracted 5,847 likely cognate pairs, 844 (or 14.43%) of which had spelling differences, which we used to train a transliteration system. The highest tuning BLEU was 95.18% (for maximum phrase size and LM order of 10), but the baseline was 93.15%. We then re-trained the system on the combination of the training and the development datasets, and we transliterated the Malay side of the training *ml-en* bi-text, which yielded a new *$ml_{in}$-en* training bi-text.

## 7. Experiments and Evaluation

Below we describe our baseline system, and we further perform various experiments to assess the similarity between the original (Indonesian and Spanish) and the auxiliary languages (Malay and Portuguese). We then improve Indonesian→English and Spanish→English SMT using Malay and Portuguese, respectively, as auxiliary languages.

We also take a closer look at improving Spanish→English SMT, performing a number of additional experiments. First, we try using an additional language that is more dissimilar to Spanish, substituting Portuguese with Italian. Second, we experiment with two auxiliary languages simultaneously: Portuguese and Italian. Finally, we combine our method with two orthogonal rivaling approaches: (1) using cognates between the source and the target language (Kondrak et al., 2003), and (2) source-language side paraphrasing with a pivot language (Callison-Burch et al., 2006).

### 7.1 Baseline SMT System

In the baseline, we used the following setup: We first tokenized and lowercased both sides of the training bi-text. We then built separate directed word alignments for English→X and X→English (X∈{Indonesian, Spanish}) using IBM model 4 (Brown, Della Pietra, Della Pietra, & Mercer, 1993), we combined them using the *intersect+grow heuristic* (Koehn et al., 2007), and we extracted phrase pairs of maximum length seven. We thus obtained a phrase table where each phrase pair is associated with the five standard parameters: forward and reverse phrase translation probabilities, forward and reverse lexical translation probabilities, and phrase penalty. We then trained a log-linear model using standard SMT feature functions: trigram language model probability, word penalty, distance-based[10] distortion cost, and the parameters from the phrase table.

---

10. We also tried lexicalized reordering (Koehn, Axelrod, Mayne, Callison-Burch, Osborne, & Talbot, 2005). While it yielded higher absolute BLEU scores, the relative improvement for a sample of our experiments was very similar to that achieved with distance-based re-ordering.

We set all weights by optimizing BLEU (Papineni, Roukos, Ward, & Zhu, 2002) using MERT on a separate development set of 2,000 sentences (Indonesian or Spanish), and we used them in a beam search decoder (Koehn et al., 2007) to translate 2,000 test sentences (Indonesian or Spanish) into English. Finally, we detokenized the output, and we evaluated it against a lowercased gold standard using BLEU.

## 7.2 Cross-lingual Translation Experiments

| # | Train | Dev | Test | LM | 10K | 20K | 40K | 80K | 160K |
|---|-------|-----|------|-----|-----|-----|-----|-----|------|
| 1 | ml-en | ml-en | ml-en | en$_{ml}$ | 44.93 | 46.98 | 47.15 | 48.04 | 49.01 |
| 2 | ml$_{in}$-en | ml-en | ml-en | en$_{ml}$ | 38.99 | 40.96 | 41.02 | 41.88 | 42.81 |
| 3 | ml-en | ml-en | **in-en** | en$_{ml}$ | 13.69 | 14.58 | 14.76 | 15.12 | 15.84 |
| 4 | ml-en | **in-en** | **in-en** | en$_{ml}$ | 13.98 | 14.75 | 14.91 | 15.51 | 16.27 |
| 5 | ml-en | **in-en** | **in-en** | **en$_{in}$** | 15.56 | 16.38 | 16.52 | 17.04 | 17.90 |
| 6 | ml$_{in}$-en | **in-en** | **in-en** | **en$_{in}$** | 16.44 | 17.36 | 17.62 | 18.14 | 19.15 |

Table 1: **Malay-Indonesian cross-lingual SMT experiments: training on Malay and testing on Indonesian for different number of training *ml-en* sentence pairs**. Columns 2-5 present the bi-texts used for training, development, and testing, and the monolingual data used to train the English language model. The following columns show the resulting BLEU (in %) for different numbers of *ml-en* training sentence pairs. Lines 1-2 show the results when training, tuning, and testing on Malay, followed by lines 3-6 on results for training on Malay but testing on Indonesian. Here ml$_{in}$ stands for Malay transliterated as Indonesian, and en$_{ml}$ and en$_{in}$ refer to the English side of the *ml-en* and *in-en* bi-text, respectively.

Here, we study the similarity between the original and the auxiliary languages.

First, we measured the vocabulary overlap between the original and the auxiliary languages. For Spanish and Portuguese, this was feasible since our training *pt-en* and *es-en* bi-texts are from the same time span in the *Europarl* corpus and their English sides largely overlap. We found 110,053 Portuguese and 121,444 Spanish word types in the *pt-en* and *es-en* bi-texts, respectively, and 44,461 of them were identical, which means that 40.40% of the Spanish word types are present on the Portuguese side of the *pt-en* bi-text. Unfortunately, we could not directly measure the vocabulary overlap between Malay and Indonesian in the same way since the English sides of the *in-en* and *ml-en* bi-texts do not overlap in content.

Second, following the general experimental setup of the baseline system, we performed cross-lingual experiments, training on one language pair and testing on another one, in order to assess the cross-lingual similarity for Indonesian-Malay and Spanish-Portuguese, and the potential of combining their corresponding training bi-texts. The results are shown in Tables 1 and 2. As we can see, this cross-lingual evaluation – training on *ml-en* (*pt-en*) instead of *in-en* (*es-en*), and testing on *in* (*es*) text – yielded a huge decrease in BLEU compared to the baseline: three times (for Malay) to five times (for Spanish) – even for very large training datasets, and even when a proper English LM and development dataset were used: compare line 1 to lines 3-5 in Table 1, and line 1 to lines 3-4 in Table 2.

| # | Train | Dev | Test | LM | 10K | 20K | 40K | 80K | 160K | 320K | 640K | 1.23M |
|---|-------|-----|------|-----|-----|-----|-----|-----|------|------|------|-------|
| 1 | pt-en | pt-en | pt-en | en$_{es:pt}$ | 21.28 | 23.11 | 24.43 | 25.72 | 26.43 | 27.10 | 27.78 | 27.96 |
| 2 | pt$_{es}$-en | pt-en | pt-en | en$_{es:pt}$ | 10.91 | 11.56 | 12.16 | 12.50 | 12.83 | 13.27 | 13.48 | 13.71 |
| 3 | pt-en | pt-en | **es-en** | **en**$_{es:pt}$ | 4.40 | 4.77 | 4.57 | 5.02 | 4.99 | 5.32 | 5.08 | 5.34 |
| 4 | pt-en | **es-en** | **es-en** | **en**$_{es:pt}$ | 4.91 | 5.12 | 5.64 | 5.82 | 6.35 | 6.87 | 6.44 | 7.10 |
| 5 | pt$_{es}$-en | **es-en** | **es-en** | **en**$_{es:pt}$ | 8.18 | 9.03 | 9.97 | 10.66 | 11.35 | 12.26 | 12.69 | 13.79 |
| 6 | es-en | es-en | es-en | **en**$_{es:pt}$ | 22.87 | 24.71 | 25.80 | 27.08 | 27.90 | 28.46 | 29.51 | 29.90 |
| 7 | es-en | es-en | **pt-en** | **en**$_{es:pt}$ | 2.99 | 3.14 | 3.33 | 3.54 | 3.37 | 3.94 | 4.18 | 3.99 |

Table 2: **Portuguese-Spanish cross-lingual SMT experiments: training on Portuguese and testing on Spanish for different number of training** *pt-en* **sentence pairs**. Lines 1-2 show the results when training, tuning, and testing on Portuguese, lines 3-5 are for training on Portuguese but testing on Spanish, and lines 6-7 are for training on Spanish and testing on Spanish or Portuguese. Columns 2-5 present the bi-texts used for training, development, and testing, and the monolingual data used to train the English language model. The following columns show the resulting BLEU (in %) for different numbers of training sentence pairs. Here pt$_{es}$ stands for Portuguese transliterated as Spanish. The English LMs for *pt-en* and *es-en* are the same (marked as en$_{es:pt}$).

For Portuguese-Spanish, we further show results in the other direction, training on Spanish and testing on Portuguese: compare line 6 to line 7 in Table 2. The results show a comparable, slightly larger, drop in BLEU for that direction. We did not carry out reverse direction experiments for Malay-Indonesian since we do not have enough parallel *in-en* data.

Third, we experimented with transliteration changing Malay to look like Indonesian and Portuguese to look like Spanish. This caused the BLEU score to double for Spanish (compare line 5 to lines 3-4 in Table 2, but improved far less for Indonesian (compare line 6 to lines 3-5 in Table 1). Training on the transliterated data and testing on Malay/Portuguese yielded about 10% relative decrease for Malay but 50% for Portuguese[11]: compare line 1 to line 2 in Tables 1 and 2. Thus, unlike Spanish and Portuguese, we found far less systematic spelling variations between Malay and Indonesian. A closer inspection confirmed this: many extracted likely Malay-Indonesian cognate pairs with spelling differences were in fact forms of a word existing in both languages, e.g., *kata* and *berkata* ('to say').

One interesting result in Table 1 is that switching the language model trained on en$_{ml}$ to one trained on en$_{in}$ yields significant improvements (compare lines 4 and 5 in Table 1). This may appear striking since the former monolingual English text is about five times bigger than the latter one, yet, this smaller language model yields better results. This is due to a partial domain shift, especially, with respect to named entities: even though both texts are in English and from the same domain, they discuss events in different countries, which involve country-specific cities, companies, political parties and their leaders; a good language model should be able to prefer good English translations of such named entities.

---

11. Interestingly, as lines 2 and 5 in Table 2 show, a system trained on 1.23M transliterated *pt$_{es}$-en* sentence pairs performs equally well when translating Portuguese and Spanish *input* text: 13.71% *vs.* 13.79%.

### 7.3 Improving Indonesian→English SMT using Malay

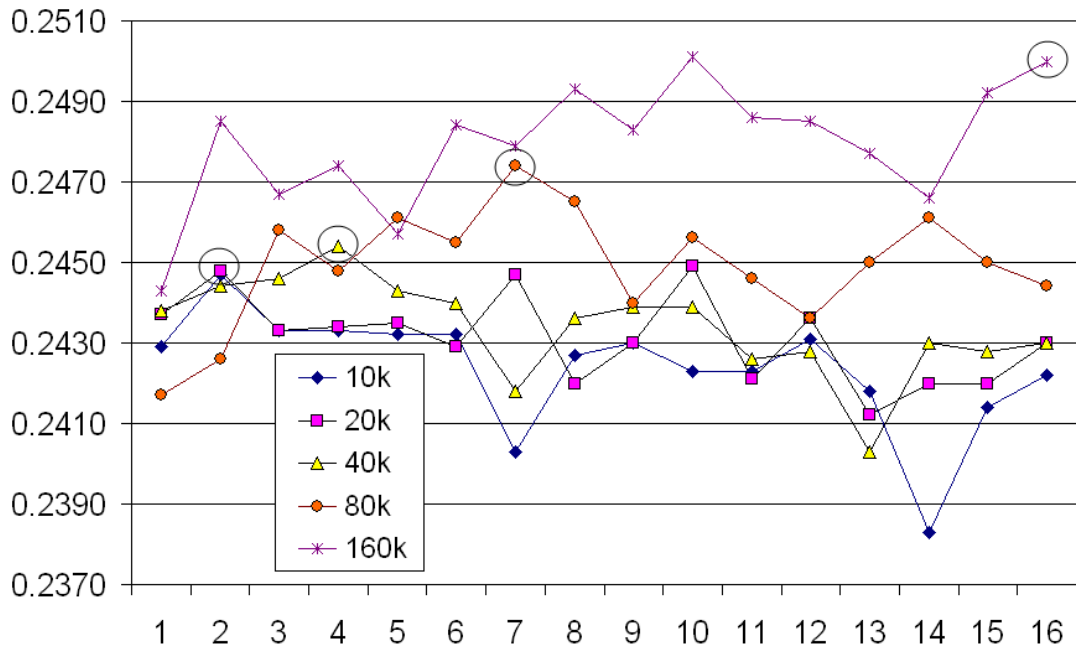

Figure 1: **Impact of $k$ on BLEU for** $cat \times k$ **for different number of extra** *ml-en*
**sentence pairs in Indonesian→English SMT.** Shown are BLEU scores for
different numbers of $k$ = 1,2,...,16 repetitions of *in-en* when concatenated to
$10000n$ pairs from *ml-en*, $n \in \{1,2,4,8,16\}$.

First, we study the impact of $k$ on **cat**$\times k$. for Indonesian→English SMT using Malay
as an additional language. We tried all values of $k$ such that $1 \leq k \leq 16$ with $10000n$ extra
*ml-en* sentence pairs, $n \in \{1,2,4,8,16\}$. As we can see in Figure 1, the highest BLEU scores
are achieved for $(n;k) \in \{(1;2),(2;2),(4;4),(8;7),(16;16)\}$, *i.e.*, when $k \approx n$. Thus, in order to
limit the search space, we used this relationship between $k$ and $n$ in our experiments (also
for Portuguese and Spanish). We should note that there is a lot of fluctuation in the results
in Figure 1, which is probably due to the small sizes of the training corpora. Given this
fluctuation, the results should not be over-interpreted, e.g., it may be just by chance that
there are peaks in the different curves at just the "right" places. Still, the overall tendency
is visible: we need to keep the balance between the original and the auxiliary bi-texts.

Tables 3 and 4 show the results for experiments on improving Indonesian→English SMT
using 10K, 20K, ..., 160K additional pairs of *ml-en* parallel sentences. Table 3 compares
the performance of our approach to the baseline and to the three concatenation meth-
ods described in Section 4.1: **cat**$\times 1$, **cat**$\times k$, and **cat**$\times k$**:align**, while Table 4 compares
the performance of our approach to various alternative ways of combining two phrase ta-
bles, namely, using alternative decoding paths, phrase table interpolation, and phrase table
merging, which were introduced in Section 4.2.

| in-en | ml-en | Baseline | cat×1 | cat×k | cat×k:align | Our approach |
|---|---|---|---|---|---|---|
| 28.4K | 10K | $23.80^<$ | $24.29^<$ | $24.29^<_{(1)}$ | $24.01^<_{(1)}$ | $^<\mathbf{24.51}_{(2;1)}$ (+0.72) |
| 28.4K | 20K | $23.80^<$ | $24.37^<$ | $^\leq 24.48_{(2)}$ | $^<24.35^<_{(2)}$ | $^<\mathbf{24.70}_{(2;2)}$ (+0.90) |
| 28.4K | 40K | $23.80^<$ | $24.38^\leq$ | $^\leq 24.54_{(4)}$ | $^<24.39^<_{(4)}$ | $^<\mathbf{24.73}_{(4;2)}$ (+0.93) |
| 28.4K | 80K | $23.80^<$ | $24.17^<$ | $^\leq 24.65^<_{(8)}$ | $24.18^<_{(8)}$ | $^<\mathbf{24.97}_{(8;3)}$ (+1.17) |
| 28.4K | 160K | $23.80^<$ | $^\leq 24.43^<$ | $^<25.00_{(16)}$ | $^\leq 24.27^<_{(16)}$ | $^<\mathbf{25.15}_{(16;3)}$ (+1.35) |

Table 3: **Improving Indonesian→English SMT using different numbers of additional Malay-English sentence pairs (varying the amount of *additional* data): concatenations, repetitions, truncations, and our approach.** The baseline is for 28,383 *in-en* sentence pairs only. Shown are the BLEU scores (in %) for different approaches. A subscript shows the best parameter value(s) found on the development set and used on the test set to produce the given result: the first value is the number of repetitions of the original bi-text while the second value, if any, is the number of extra features added to the phrase table. The BLEU scores that are statistically significantly better than the baseline/our approach are marked on the left/right side by $^<$ (for $p < 0.01$) or $^\leq$ (for $p < 0.05$).

| in-en | ml-en | Baseline | Two Tables | Interpolation | Merge | Our approach |
|---|---|---|---|---|---|---|
| 28.4K | 10K | $23.80^<$ | $^\geq 23.79^<$ | $23.89^<_{(.9)}$ | $23.97^<_{(3)}$ | $^<\mathbf{24.51}_{(2;1)}$ (+0.72) |
| 28.4K | 20K | $23.80^<$ | $24.24^<$ | $24.22^<_{(.8)}$ | $^\leq 24.46^<_{(3)}$ | $^<\mathbf{24.70}_{(2;2)}$ (+0.90) |
| 28.4K | 40K | $23.80^<$ | $24.27^<$ | $24.27^<_{(.8)}$ | $24.43^<_{(3)}$ | $^<\mathbf{24.73}_{(4;2)}$ (+0.93) |
| 28.4K | 80K | $23.80^<$ | $24.11^<$ | $^\leq 24.46^<_{(.8)}$ | $^<24.67_{(3)}$ | $^<\mathbf{24.97}_{(8;3)}$ (+1.17) |
| 28.4K | 160K | $23.80^<$ | $^<24.58^<$ | $^<24.58^<_{(.8)}$ | $^<24.79^\leq_{(3)}$ | $^<\mathbf{25.15}_{(16;3)}$ (+1.35) |

Table 4: **Improving Indonesian→English SMT using different numbers of additional Malay-English sentence pairs (varying the amount of *additional* data): comparing our approach to various alternatives.** The baseline is for 28,383 *in-en* sentence pairs only. Shown are the BLEU scores (in %) for different approaches. A subscript shows the best parameter value(s) found on the development set and used on the test set to produce the given result: for merging methods, the first value is the number of repetitions of the original bi-text while the second value, if any, is the number of extra features added to the phrase table; for interpolation, we show the weight of the phrase pairs from *in-en*. The BLEU scores that are statistically significantly better than the baseline/our approach are marked on the left/right side by $^<$ (for $p < 0.01$) or $^\leq$ (for $p < 0.05$).

Several interesting general observations about Tables 3 and 4 can be made. First, using more additional Indonesian-English sentences yields better results. Second, with one exception, all experiments yield improvements over the baseline. Third, the improvements are always statistically significant for **our approach**, according to Collins, Koehn, and Kučerová's (2005) sign test.

Overall, among the different bi-text combination strategies, **our approach** performs best, followed by **cat×k**, **merge**, and **interpolation**, which are very close in performance; these three strategies are the only ones to consistently yield higher BLEU as the number of additional *ml-en* sentence pairs grows. Methods like **cat×1**, **cat×k:align**, and **two-tables** are somewhat inconsistent in that respect. The latter method performs worst and is the only one to go below the baseline (for 10K *ml-en* sentence pairs).

One possible reason for the relatively bad performance of **two-tables** could be that it has to tune more weights compared to the other models: each phrase table has its own feature weights, which means five additional features. It is well known that MERT cannot handle too many features (Chiang, Knight, & Wang, 2009; Hopkins & May, 2011), and we believe this is our case as it takes 30–35 iterations to finish, while the other methods normally only need 7–8 iterations. A closer look at MERT revealed two further issues: (1) The *n*-best list had many identical translations, *i.e.*, spurious ambiguity became an even bigger problem. (2) In MERT, identical translations had different feature values, which could have confused the optimization. We believe these problems were caused by the fact that often two identical translations would be found that use the same phrases but from the different tables and thus with different scores.

Note also the high values of the interpolation parameter $\alpha$ in Table 4: 0.8–0.9. They indicate that the original bi-text needs to be weighted higher than the auxiliary one, thus supporting the need for balanced concatenations with repetitions of the original bi-text, and indirectly explaining why **cat×k** performs better than **cat×1** in Table 3.

## 7.4 Improving Spanish→English SMT using Portuguese

Next, we experiment with using Portuguese to improve Spanish→English SMT.

The results are shown in Tables 5 and 6. Overall, they are consistent with those for Indonesian→English SMT using the additional Malay-English bi-text (shown in Tables 3 and 4 above). We can further observe that, as the size of the original bi-text increases, the gain in BLEU decreases, which is to be expected. Note also that here transliteration is very important: it doubles the absolute gain in BLEU achieved by our method.

Table 7 compares the performance of our technique for 160K *vs.* 1.23M *additional pt-en* parallel sentence pairs, with and without transliteration for training bi-texts with different numbers of parallel *es-en* sentence pairs (10K, 20K, …, 320K). The table shows the importance of transliteration, which is responsible for about half of the improvement over the baseline brought by our method. In fact, for small original *es-en* bi-texts (10K, 20K, 40K), using 160K of transliterated additional *pt-en* sentence pairs works better than using 1.23M additional non-transliterated *pt-en* sentence pairs (which is eight times bigger). For example, given 10K of original training *es-en* sentence pairs, going from 160K to 1.23M additional *pt-en* sentence pairs improves BLEU by 0.25% only (from 23.98% to 24.23%), while using 160K of transliterated *pt-en* data yields an improvement of 1.75% (from 23.98% to 25.73%). The impact of transliteration should not be surprising: we have already seen it in Table 2, where, comparing lines 4 and 5, we can see that transliterating Portuguese to look like Spanish effectively doubles the BLEU score: from 4.91% to 8.18% for 10K, and from 7.10% to 13.79% for 1.23M parallel training sentence pairs.

| es-en | pt-en | Translit. | Baseline | cat×1 | cat×k | cat×k:align | Our method |
|---|---|---|---|---|---|---|---|
| 10K | 160K | no | $22.87^<$ | $^<23.54^<$ | $^<23.83^<_{(16)}$ | $22.93^<_{(16)}$ | $^<\mathbf{23.98}_{(16;3)}$ (+1.11) |
| | | yes | $22.87^<$ | $^<25.26$ | $^<25.42_{(16)}$ | $^<23.31^<_{(16)}$ | $^<\mathbf{25.73}_{(16;3)}$ (+2.86) |
| 20K | 160K | no | $24.71^<$ | $^<25.19^<$ | $^<25.29^<_{(8)}$ | $24.91^<_{(8)}$ | $^<\mathbf{25.65}_{(8;2)}$ (+0.94) |
| | | yes | $24.71^<$ | $^<26.16^\le$ | $^<26.18^\le_{(8)}$ | $24.88^<_{(8)}$ | $^<\mathbf{26.36}_{(8;3)}$ (+1.65) |
| 40K | 160K | no | $25.80^<$ | $26.24^<$ | $25.92^<_{(4)}$ | $25.99^<_{(4)}$ | $^<\mathbf{26.49}_{(4;2)}$ (+0.69) |
| | | yes | $25.80^<$ | $^<26.78$ | $^<26.93_{(4)}$ | $25.88^<_{(4)}$ | $^<\mathbf{26.95}_{(4;3)}$ (+1.15) |
| 80K | 160K | no | $27.08^\le$ | $27.23$ | $27.09^<_{(2)}$ | $27.01^<_{(2)}$ | $^\le\mathbf{27.30}_{(2;2)}$ (+0.22) |
| | | yes | $27.08^<$ | $27.26^<$ | $^\le27.53^<_{(2)}$ | $27.09^<_{(2)}$ | $^<\mathbf{27.49}_{(2;3)}$ (+0.41) |
| 160K | 160K | no | $27.90$ | $27.83^<$ | $27.83^<_{(1)}$ | $27.94_{(1)}$ | $\mathbf{28.05}_{(1;3)}$ (+0.15) |
| | | yes | $27.90$ | $^\le28.14$ | $^\le28.14_{(1)}$ | $28.06_{(1)}$ | $\mathbf{28.16}_{(1;2)}$ (+0.26) |

Table 5: **Improving Spanish→English SMT using 160K additional Portuguese-English sentence pairs (varying the amount of *original* data): concatenations, repetitions, truncations, and our method.** The first column contains the number of original (*es-en*) sentence pairs. Column 3 shows whether transliteration was used; the following columns list the BLEU scores (in %) for different methods. A subscript shows the best parameter value(s) found on the development set and used on the test set to produce the given result: the first value is the number of repetitions of the original bi-text while the second value, if any, is the number of extra features added to the phrase table. The BLEU scores that are statistically significantly better than the baseline/our method are marked on the left/right side by $^<$ (for $p < 0.01$) or $^\le$ (for $p < 0.05$).

Note that the impact of transliteration diminishes as the size of the *es-en* bi-text grows. This should not be surprising: as the size of the good original *es-en* bi-text grows, there is less and less to be learned from the additional *pt-en* bi-text, regardless of whether with or without transliteration.

### 7.5 Improving Spanish→English SMT Using Italian

Here, we experiment with Italian as an auxiliary language for improving Spanish→English phrase-based SMT. Figure 2 shows the results when using Italian and Portuguese as auxiliary languages in our method with transliteration. We can see a major consistent drop in BLEU score when using Italian instead of Portuguese. For example, for 10K *es-en* sentence pairs and 160K additional *pt-en/it-en* sentence pairs, there is an absolute drop in BLEU by about 0.9%: we have 25.73% *vs.* 24.82%, respectively. Moreover, for 160K original *es-en* sentence pairs, our method goes slightly below the baseline (by -0.05) when using *it-en* while there is a small improvement (by +0.26) for *pt-en*.

Still, Figure 2 shows that Italian, which is more dissimilar to Spanish than Portuguese, is useful as an auxiliary language for smaller sizes of the original *es-en* training bi-text. Thus, we can conclude that while the degree of similarity between the auxiliary and the source language does matter, more dissimilar languages are still potentially useful as auxiliary languages.

| es-en | pt-en | Translit. | Baseline | Two tables | Interpol. | Merge | Our method |
|---|---|---|---|---|---|---|---|
| 10K | 160K | no | $22.87^<$ | $^<23.81$ | $^<23.73_{(.5)}$ | $^<23.60_{(2)}$ | $^<\mathbf{23.98}_{(16;3)}$ (+1.11) |
| | | yes | $22.87^<$ | $^<25.29^\leq$ | $^<25.22_{(.5)}$ | $^<25.16_{(2)}$ | $^<\mathbf{25.73}_{(16;3)}$ (+2.86) |
| 20K | 160K | no | $24.71^<$ | $^<25.22$ | $^\leq25.02_{(.5)}$ | $^<25.32_{(3)}$ | $^<\mathbf{25.65}_{(8;2)}$ (+0.94) |
| | | yes | $24.71^<$ | $^<26.07^\leq$ | $^<26.07_{(.7)}$ | $^<26.04_{(3)}$ | $^<\mathbf{26.36}_{(8;3)}$ (+1.65) |
| 40K | 160K | no | $25.80^<$ | $25.96^<$ | $26.15^<_{(.6)}$ | $25.99^<_{(3)}$ | $^<\mathbf{26.49}_{(4;2)}$ (+0.69) |
| | | yes | $25.80^<$ | $^<26.68$ | $^<26.43_{(.7)}$ | $^<26.64_{(3)}$ | $^<\mathbf{26.95}_{(4;3)}$ (+1.15) |
| 80K | 160K | no | $27.08^\leq$ | $^\geq26.89^<$ | $27.04_{(.8)}^<$ | $27.02_{(3)}^<$ | $^<\mathbf{27.30}_{(2;2)}$ (+0.22) |
| | | yes | $27.08^<$ | $27.20^<$ | $27.42_{(.5)}$ | $27.29^<_{(3)}$ | $^<\mathbf{27.49}_{(2;3)}$ (+0.41) |
| 160K | 160K | no | $27.90$ | $27.99$ | $27.72_{(.5)}$ | $27.95_{(2)}$ | $\mathbf{28.05}_{(1;3)}$ (+0.15) |
| | | yes | $27.90$ | $28.11$ | $^\leq28.13_{(.6)}$ | $^\leq28.17_{(2)}$ | $\mathbf{28.16}_{(1;2)}$ (+0.26) |

Table 6: **Improving Spanish→English SMT using 160K additional Portuguese-English sentence pairs (varying the amount of *original* data): comparing our method to various alternatives.** The first column contains the number of original (*es-en*) sentence pairs. Column 3 shows whether transliteration was used; the following columns list the BLEU scores (in %) for different methods. A subscript shows the best parameter value(s) found on the development set and used on the test set to produce the given result: for merging methods, the first value is the number of repetitions of the original bi-text while the second value, if any, is the number of extra features added to the phrase table; for interpolation, we show the weight of the phrase pairs from *in-en*. The BLEU scores that are statistically significantly better than the baseline/our method are marked on the left/right side by $^<$ (for $p < 0.01$) or $^\leq$ (for $p < 0.05$).

## 7.6 Improving Spanish→English SMT Using Both Portuguese and Italian

After having seen that both Portuguese and Italian are useful as auxiliary languages, we tried to use them both together. The experiments were carried out in the same way as when we used a single auxiliary language, except that now we had to double the usual number of repetitions $k$ of the original bi-text so that the auxiliary bi-texts do not dominate it for **cat×k:align**. For example, for 10K original *es-en* training sentence pairs and 160K auxiliary *pt-en* and 160K *it-en* sentence pairs, we need to include 32 copies of the original bi-text instead of 16, as we were doing before.

The results for the combination are shown in Figure 2. Comparing them to the results when using *pt-en* data only, we can see that there is a small but consistent improvement. For example, for 10K original *es-en* sentence pairs, 160K additional *pt-en* and 160K additional *it-en* sentence pairs, there is an absolute increase in BLEU scores by 0.18%: from 25.73% to 25.91%. The size of the absolute improvement when using 20K, 40K, 80K, and 160K additional *pt-en* and *it-en* sentence pairs is comparable: about 0.10-0.20% on average.

Thus, there are potential gains when using multiple auxiliary languages simultaneously.

| System | 10K | 20K | 40K | 80K | 160K | 320K |
|---|---|---|---|---|---|---|
| baseline | 22.87 | 24.71 | 25.80 | 27.08 | 27.90 | 28.46 |
| our method: **160K** *pt-en* pairs | 23.98* | 25.65* | 26.49* | 27.30$^\diamond$ | 28.05 | 28.52 |
| – improvement | **+1.11*** | **+0.94*** | **+0.69*** | **+0.22$^\diamond$** | **+0.15** | **+0.06** |
| our method: **1.23M** *pt-en* pairs | 24.23* | 25.70* | 26.78* | 27.49 | 28.22$^\diamond$ | 28.58 |
| – improvement | **+1.36*** | **+0.99*** | **+0.98*** | **+0.41** | **+0.32$^\diamond$** | **+0.12** |
| our method: **160K** *pt-en*, **translit.** | 25.73* | 26.36* | 26.95* | 27.49* | 28.16 | 28.43 |
| – improvement | **+2.86*** | **+1.65*** | **+1.15*** | **+0.41*** | **+0.26** | **-0.03** |
| our method: **1.23M** *pt-en*, **translit.** | 26.24* | 26.82* | 27.47* | 27.85* | 28.50* | 28.70 |
| – improvement | **+3.37*** | **+2.11*** | **+1.67*** | **+0.77*** | **+0.60*** | **+0.24** |

Table 7: **Spanish→English: testing our method using 160K *vs.* 1.23M additional *pt-en* sentence pairs, with and without transliteration.** Shown are BLEU scores (in %) and absolute improvement over the baseline for training bi-texts with different numbers of parallel *es-en* sentence pairs (10K, 20K, ..., 320K) and a fixed number of *additional pt-en* sentence pairs: 160K and 1.23M. All statistically significant improvements over the baseline are marked with a * (for $p < 0.01$) and with a $^\diamond$ (for $p < 0.05$).

### 7.7 Combining Our Method with the Cognate Extraction Technique of Kondrak et al. (2003)

Next, we combined our method with the cognate extraction technique of Kondrak et al. (2003), where pairs of likely cognates are extracted from the original training bi-text and then added to that bi-text as additional 1-word-to-1-word sentence pairs.

| System | 10K | 20K | 40K | 80K | 160K | 320K |
|---|---|---|---|---|---|---|
| baseline | 22.87 | 24.71 | 25.80 | 27.08 | 27.90 | 28.46 |
| cognates | 23.50 | 25.22 | 26.31 | 27.38 | 28.10 | 28.74 |
| – improvement (baseline) | **+0.63*** | **+0.51*** | **+0.51*** | **+0.30$^\diamond$** | **+0.20** | **+0.28$^\diamond$** |
| our (**1.23M** *pt-en pt-en*) + cognates | 24.55 | 25.98 | 26.73 | 27.67 | 28.33 | 28.90 |
| – improvement (baseline) | **+1.68*** | **+1.27*** | **+0.93*** | **+0.59*** | **+0.43*** | **+0.44*** |
| – improvement (our: **1.23M** *pt-en*) | *+0.32$^\diamond$* | *+0.28$^\diamond$* | *-0.05* | *+0.18* | *+0.11* | *+0.32$^\diamond$* |
| our (**1.23M** *pt-en*, **transl.**) + cognates | 26.35 | 26.78 | 27.34 | 27.79 | 28.50 | 28.68 |
| – improvement (baseline) | **+3.48*** | **+2.07*** | **+1.54*** | **+0.71*** | **+0.60*** | **+0.22$^\diamond$** |
| – improvement (our: **1.23M**, **transl.**) | *+0.11* | *-0.04* | *-0.13* | *-0.06* | *+0.00* | *-0.02* |

Table 8: **Spanish→English: combining our method with the cognate extraction technique of Kondrak et al. (2003).** Shown are BLEU scores (in %) and absolute improvements (over the baseline and over our method) for training bi-texts with different numbers of parallel *es-en* sentence pairs (10K, 20K, ..., 320K) and fixed number of *additional pt-en* sentence pairs (1.23M), with and without transliteration. The statistically significant improvements are marked with a * (for $p < 0.01$) and with a $^\diamond$ (for $p < 0.05$).

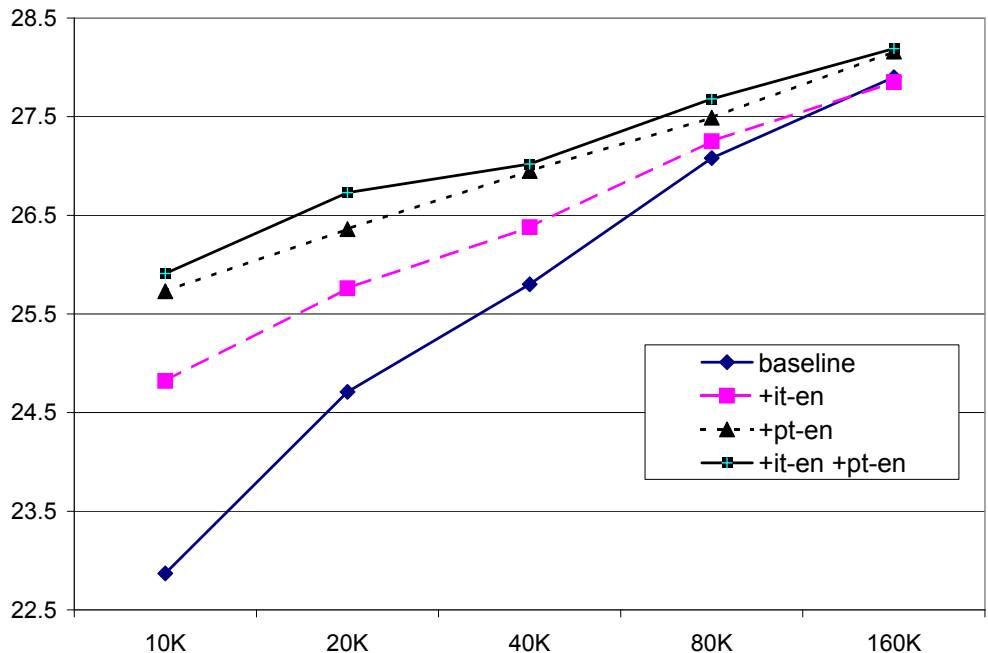

Figure 2: **Improving Spanish→English SMT using 160K Italian-English and 160K Portuguese-English additional sentence pairs (varying the amount of *original* data) and transliteration.**

The results for adding these cognates to the training *es-en* bi-text, *i.e.*, for our re-implementation of their algorithm, are shown in the top lines of Table 8. We can see an absolute improvement of 0.5% BLEU for *es-en* of size up to 40K, and the improvement is statistically significant.

Next, we combined our method with the cognate extraction method as follows: first, we augmented the original *es-en* bi-text with cognate pairs, and then we used this augmented bi-text instead of *es-en* in our method. Table 8 shows the results for the combination of our method with cognate extraction (BLEU scores in % and absolute improvements over the baseline and over our method) for training bi-texts with different numbers of parallel *es-en* sentence pairs (10K, 20K, ..., 320K) and fixed number of *additional pt-en* sentence pairs (1.23M), with and without transliteration. As we can see, it is worth combining our method with the cognate extraction technique of Kondrak et al. (2003) for small original *es-en* datasets, e.g., 10K or 20K (in which cases statistically significant improvements occur over using our method only), but only when our method does not use transliteration.

We found it interesting that combining our method with the cognate extraction technique of Kondrak et al. (2003) does not help so much when we used transliteration compared to when we do not use it. Thus, we further analyzed the case of 10K *es-en* sentence pairs and 1.23M *pt-en* pairs. The cognate extraction technique yielded 25,362 Spanish-English likely cognate pairs, including 10,611 unique Spanish words. The Portuguese side of the 1.23M *pt-en* data contained only 14 or 0.13% of these 10,611 unique Spanish words. Thus, the information that these Spanish-English likely cognates provide for word alignments and phrase pairs is clearly complementary to what the *pt-en* bi-text gives. However, after transliteration, the source side of the 1.23M $pt_{es}$-*en* bi-text contained 8,867 or 83.56% of the 10,611 unique Spanish words in the Spanish-English likely cognate pairs. This drastic jump means that the Spanish-English likely cognate pairs have little to add on top of what $pt_{es}$-*en* already provides, and explains the lack of improvement when combined with our method when transliteration is used.

## 7.8 Combining Our Method with the Phrase Table Pivoting Technique of Callison-Burch et al. (2006)

Finally, we combined our method with the phrase table pivoting technique of Callison-Burch et al. (2006) since they are orthogonal.

First, we tried to reproduce the phrase table pivoting experiments of Callison-Burch et al. (2006), which turned out to be complicated (even though we used their original code to do the pivoting) because of various differences in our experimental setups: (1) we used Moses instead of Pharaoh for translation; (2) we used IRSTLM instead of SRILM for language modeling; (3) we used different tokenization; (4) we used a maximum phrase length of up to seven instead of ten; (5) we created our training/dev/test dataset out of Europarl v.3, which is different from the version of the Europarl corpus that was available in 2006 (which also implies a different baseline, *etc.*).

The results are shown in Table 9. The bottom three lines show the results reported by Callison-Burch et al. (2006), while the top three lines report the BLEU scores for our reproduction of their experiments, in which about 1.3M pairs were used for each of eight additional pivot languages: Danish, Dutch, Finnish, French, German, Italian, Portuguese, and Swedish. While our BLEU scores are lower, they are good enough for studying the potential of combining the two methods.

The combination was carried in the following way: after we had built the final merged phrase table for our method, we paraphrased its source side through pivoting using the method of Callison-Burch et al. (2006). The middle lines of the table show the BLEU scores (in %) of the combined method and absolute improvements (over the baseline and over our method) for training bi-texts with different numbers of parallel *es-en* sentence pairs (10K, 20K, ..., 320K) and fixed amount of *additional pt-en* pairs (160K and 1.23M pairs), with and without transliteration.

The results show that it is worth combining our method with phrase table pivoting for small *es-en* datasets, e.g., 10K or 20K (in which cases, statistically significant improvements occur over using our method only), but only when our method does not use transliteration, as was the case for the cognate extraction technique of Kondrak et al. (2003).

| Our Experiments | 10K | 20K | 40K | 80K | 160K | 320K |
|---|---|---|---|---|---|---|
| baseline | 22.87 | 24.71 | 25.80 | 27.08 | 27.90 | 28.46 |
| Pivoting (+8 pairs × ∼1.3M) | 23.33 | 24.88 | 26.10 | 27.06 | 28.09 | 28.49 |
| – improvement (baseline) | **+0.46*** | **+0.17** | **+0.30$^\diamond$** | **-0.02** | **+0.19** | **+0.03** |
| our (**160K** *pt-en*) + pivoting | 24.32 | 25.95 | 26.70 | 27.36 | 28.02 | 28.56 |
| – improvement (over the baseline) | **+1.45*** | **+1.24*** | **+0.90*** | **+0.28$^\diamond$** | **+0.12** | **+0.10** |
| – improvement (over our method) | *+0.34$^\diamond$* | *+0.30$^\diamond$* | *+0.21* | *+0.06* | *-0.03* | *+0.04* |
| our (**1.23M** *pt-en*) + pivoting | 24.64 | 26.18 | 26.87 | 27.60 | 28.35 | 28.69 |
| – improvement (over the baseline) | **+1.77*** | **+1.47*** | **+1.07*** | **+0.52*** | **+0.45*** | **+0.23** |
| – improvement (over our method) | *+0.41*** | *+0.48*** | *+0.09* | *+0.11* | *+0.13* | *+0.11* |
| our (**160K** *pt-en*, **transl.**) + pivoting | 25.82 | 26.49 | 27.06 | 27.51 | 28.35 | 28.58 |
| – improvement (over the baseline) | **+2.95*** | **+1.78*** | **+1.26*** | **+0.43*** | **+0.45*** | **+0.12** |
| – improvement (over our method) | *+0.09* | *+0.13* | *+0.11* | *+0.02* | *+0.19* | *+0.15* |
| our (**1.23M** *pt-en*, **transl.**) + pivoting | 26.39 | 27.01 | 27.53 | 27.77 | 28.58 | 28.66 |
| – improvement (over the baseline) | **+3.52*** | **+2.30*** | **+1.73*** | **+0.69*** | **+0.68*** | **+0.20** |
| – improvement (over our method) | *+0.15* | *+0.19* | *+0.06* | *-0.08* | *+0.08* | *-0.04* |
| **Callison-Burch et al. (2006)** | | | | | | |
| *baseline* | 22.6 | 25.0 | 26.5 | 26.5 | 28.7 | 30.0 |
| *Pivoting (+8 pairs × ∼1.3M)* | 23.3 | 26.0 | 27.2 | 28.0 | 29.0 | 30.0 |
| *– improvement (over the baseline)* | **+0.7** | **+1.0** | **+0.7** | **+1.5** | **+0.3** | **+0.0** |

Table 9: **Spanish→English: combining our method with the phrase table pivoting technique of Callison-Burch et al. (2006).** Shown are BLEU scores (in %) and absolute improvements (over the baseline and over our method) for training bi-texts with different numbers of parallel *es-en* sentence pairs (10K, 20K, ..., 320K) and fixed amount of *additional pt-en* pairs: (1) about 1.3M pairs for each of eight additional languages in pivoting, and (2) 160K and 1.23M pairs for one language (Portuguese) for our method (with and without transliteration). The last three lines show the results of the phrase table pivoting experiments reported in Callison-Burch et al. (2006) while the first three lines show our reproduction of these experiments. The statistically significant improvements are marked with a * (for $p < 0.01$) and with a $^\diamond$ (for $p < 0.05$).

Again, we found it interesting that pivoting does not help so much with transliteration as without it. Thus, we had a closer look at the interaction of pivoting and transliteration for 10K *es-en* sentence pairs and 160K *pt-en* pairs. In particular, we looked at the number of usable phrase pairs with respect to the test data, *i.e.*, those phrase pairs whose source side matches the test data, and we found that without pivoting, using transliteration increases this number from 657,541 to 1,863,950, *i.e.*, by 183.47%, while using pivoting and transliteration increases this number from 819,324 to 2,214,580, *i.e.*, by 170.29%. This lower relative increase in the number of usable phrases is one possible explanation of the corresponding lower increase in BLEU: which is +0.34 and +0.09 absolute, respectively, without and with transliteration.

## 8. Analysis and Discussion

Below, we perform deeper analysis of our method and of the obtained results.

### 8.1 Merging Phrase Tables

Here we compare merging **cat**×$k$:**align** and **cat**×1 (with 1-3 extra features, as was described above), to two simpler alternatives: (a) substituting **cat**×1 with *ml-en*, and (b) merging the phrase tables derived from the original bi-texts, *in-en* and *ml-en*.

We further implement and evaluate the following alternative to our method: (c) train the alignment models on the combined bi-text that consists of $k$ copies of *in-en* and one copy of *ml-en*, then truncate the alignments appropriately, and build two separate phrase tables. The first table is **cat**×$k$:**align** as in our method (built on one copy of *in-en*), and the second one is a similar phrase table that corresponds to *ml-en*. Unlike (a) above, the word alignments in that second phrase table are influenced by the $k$ copies of *in-en*. We will refer to this second phrase table as *ml*:**cat**×$k$:**align**. The motivation for trying this alternative is that it is (1) a bit simpler to implement, and (2) somewhat symmetric for both phrase tables. Yet, just like in our method, the alignment models benefit from more data, while the phrase tables remain language-specific and thus can be combined using extra features.

Table 10 compares our method (line 3) to the above-described three alternatives, (a), (b) and (c), for different numbers of training *ml-en* sentence pairs. As we can see, overall, our method (line 3) performs best, while the newly described alternative (c), shown on line 4, is ranked second. Merging phrase tables derived from the original bi-texts *in-en* and *ml-en* is worst (line 1), which can be explained by the fact that it cannot benefit from improved word alignments for the small *in-en* bi-text (unlike the other combinations, and most notably the one at line 2). However, it is not so easy to explain from this table alone why our method is better than the other two alternatives.

Thus, we looked into the phrase tables and the unknown words, for the case of 160K *ml-en* additional sentence pairs. The results are shown in Table 11, which offers a very good insight: the two good performing combinations at lines 3-4 simply have larger phrase tables compared to those on lines 1-2. More importantly, this translates into a higher number of phrase pairs that are potentially usable for translating the test sentences, *i.e.*, match the input at test time. Naturally, more translation options mean a larger search space and thus more opportunities to find better translations, which explains the better performance of the combinations at lines 3-4. The table also compares the number of unknown word types and word tokens when translating the test data. We can see that our method has the lowest number of unknowns, which can explain its good performance. On the other hand, the numbers of unknown words are comparable for the other three methods.

In summary, Table 11 shows two important factors influencing BLEU: (*i*) total and used number of phrase pairs, and (*ii*) number of unknown word types and tokens at translation time. Our method ranks best in both criteria, while the second best method at line 4 ranks second on the first factor but last on the second one (but close to the other methods). Thus, we could conclude that the impact of unknown words on BLEU is limited when the differences are small. The phrase table size seems to correlate somewhat better with BLEU, at least for the two best performing methods. Finally, comparing line 2 to lines 3-4, we can further conclude that using the *in-en* bi-text to help align the *ml-en* bi-text is beneficial.

To get even better insight, we looked at the characteristics of the tables that are being combined: (1) phrase table sizes and overlap, (2) number of distinct source phrases and overlap, and (3) average differences in the four standard scores in the merged tables for shared phrase pairs: inverse phrase translation probability $\phi(f|e)$, inverse lexical weighting $p_w(f|e)$, direct phrase translation probability $\phi(e|f)$, and direct lexical weighting $p_w(e|f)$. The results are shown in Table 12. The table shows that our method combines phrase tables that have a much higher overlap (10-50 times higher!), both in terms of number of phrase pairs and number of distinct source phrases. Moreover, the absolute differences in the scores for the shared phrase pairs are about halved (*i.e.*, they are very similar) for the two phrase tables that are combined by our method, **cat×1** and **cat×k:align**, compared to those for phrase tables combined by the three alternative approaches, as the last four columns in Table 12 show. This high similarity in the scores for **cat×1** and **cat×k:align** could be one possible explanation of their very similar performance as shown in Table 3.

Thus, our method wins by combining two tables that have already been made more similar, and thus more appropriate for combination. A key element of this is to build the second table from a concatenation of the *ml-en* and the *in-en* bi-texts, where the *in-en* bi-text has a minor influence on word alignment (it is simply much smaller), but much more influence on phrase extraction and scoring. This makes the resulting phrase table much more similar to the first phrase table (which has also been made similar to the second table, but via word alignments only), but also much bigger than if trained on *ml-en* data only (14M *vs.* 11M phrase pairs); this in turn yields a larger merged phrase table despite the higher overlap between the tables that are being merged.

| | **Merged Phrase Tables** | | 10K | 20K | 40K | 80K | 160K |
|---|---|---|---|---|---|---|---|
| 1 | *in-en* | *ml-en* | 23.97 | 24.46 | 24.43 | 24.67 | 24.79 |
| 2 | **cat×k:align** | *ml-en* | 24.11 | 24.56 | 24.54 | 24.62 | 25.02 |
| 3 | **cat×k:align** | **cat×1** | 24.51 | 24.70 | 24.73 | 24.97 | 25.15 |
| 4 | **cat×k:align** | *ml*:**cat×k:align** | 24.14 | 24.54 | 24.65 | 24.72 | 25.08 |

Table 10: **Merging phrase tables for Indonesian-English SMT: BLEU scores.** BLEU shown in % for different numbers of training *ml-en* sentence pairs.

## 8.2 Transliteration

Here we have a closer look at transliteration: studying how many words it affects and its impact on the number of unknown words (also known as OOVs, out-of-vocabulary words).

First, we look at the number of word types and word tokens that changed in the process of transliteration of the source side of the additional training bi-text. The results are shown in Table 13. We can see that transliterating Malay to Indonesian affects a very small number of words: 7.61% of the word types and 5.78% of the word tokens. This should not be surprising since the spelling differences between Malay and Indonesian are very limited, as we explained in Section 3.1. In contrast, transliterating Portuguese to Spanish changes 44.71% of the word types and 23.17% of the word tokens, which agrees with our observations in Sections 3.2 and 6. Italian is even more affected by transliteration than Portuguese: with 70.45% of the word types and 34.37% of the word tokens changed.

| | Merged Phrase Tables | | Phrase Pairs | | | Unknown | | |
|---|---|---|---|---|---|---|---|---|
| | | | Total | # Used | % Used | Types | Tokens | **BLEU** |
| 1 | *in-en* | *ml-en* | 14.23M | 1.28M | 9.02% | 1411 | 1917 | 24.79 |
| 2 | **cat×k:align** | *ml-en* | 14.07M | 1.25M | 8.88% | 1413 | 1906 | 25.02 |
| 3 | **cat×k:align** | **cat×1** | 15.83M | 1.70M | 10.71% | 1300 | 1743 | 25.15 |
| 4 | **cat×k:align** | *ml:***cat×k:align** | 14.92M | 1.37M | 9.17% | 1445 | 1933 | 25.08 |

Table 11: **Merging phrase tables derived from *in-en* and *ml-en* (160K): number of phrase pairs and unknown words.** Shown are the total number of phrase pairs in the merged phrase table and the number of phrase pairs used to decode the test data, followed by the number of unknown word types and tokens, and the BLEU score (in %).

| | Merged Phrase Tables | | Phrase Pairs | | | Source Phrases | | | Avg. Score Diff. | | | |
|---|---|---|---|---|---|---|---|---|---|---|---|---|
| | | | PT1 | PT2 | both | PT1 | PT2 | both | $\phi(f\|e)$ | $p_w(f\|e)$ | $\phi(e\|f)$ | $p_w(e\|f)$ |
| 1 | *in-en* | *ml-en* | 3.2M<br>2.23% | 11.1M<br>0.65% | 72.1K | 1.1M<br>0.91% | 7.7M<br>0.13% | 10.3K | 0.40 | 0.18 | 0.19 | 0.10 |
| 2 | **cat×k:align** | *ml-en* | 3.1M<br>2.51% | 11.1M<br>0.70% | 77.1K | 1.2M<br>0.95% | 7.7M<br>0.14% | 11.1M | 0.40 | 0.18 | 0.19 | 0.10 |
| 3 | **cat×k:align** | **cat×1** | 3.1M<br>39.39% | 14.0M<br>8.67% | 1.2M | 1.2M<br>49.40% | 8.8M<br>6.54% | 0.6M | 0.21 | 0.05 | 0.13 | 0.05 |
| 4 | **cat×k:align** | *ml:*cat×k:align | 3.1M<br>2.85% | 11.9M<br>0.73% | 87.6K | 1.2M<br>0.99% | 7.6M<br>0.15% | 11.6M | 0.40 | 0.18 | 0.18 | 0.09 |

Table 12: **Comparison of the phrase tables merged in Tables 10 and 11.** Shown are the number of phrase pairs / source phrases in each phrase table and the number/percent of them that appear in both tables. The last four columns show the average absolute differences in the four standard phrase table scores for phrase pairs that appear in both tables; these scores are inverse phrase translation probability $\phi(f|e)$, inverse lexical weighting $p_w(f|e)$, direct phrase translation probability $\phi(e|f)$, and direct lexical weighting $p_w(e|f)$.

One important reason for this higher number of changes would be that, unlike Spanish and Portuguese, Italian does not form plural for nouns and adjectives by adding an *-s* but by a vowel change. For example, the singular form of the adjective meaning *green* is *verde* in all three languages: Spanish, Portuguese, and Italian. However, its plural form differs: it is *verdes* regardless of gender in both Spanish and Portuguese, but it is *verdi* (plural masculine) or *verde* (plural feminine) in Italian. Thus, transliterating Portuguese to Spanish would leave *verdes* intact, but for Italian, changes would be needed. Given the frequency of the use of plural for nouns and adjectives, we can expect many more differences for Italian–Spanish than for Portuguese–Spanish. Overall, the small number of word types/tokens changed explains why transliteration was of limited use for Malay but so important for Spanish and Italian.

| | Transliteration | **Word Types** | | **Word Tokens** | |
|---|---|---|---|---|---|
| | | Changed | Total | Changed | Total |
| 1 | Malay→Indonesian | 8,259 | 108,595 | 316,444 | 5,472,372 |
| | | 7.61% | | 5.78% | |
| 2 | Portuguese→Spanish | 52,303 | 116,989 | 8,315,835 | 35,889,877 |
| | | 44.71% | | 23.17% | |
| 3 | Italian→Spanish | 88,767 | 126,005 | 14,962,680 | 43,530,246 |
| | | 70.45% | | 34.37% | |

Table 13: **Transliteration: number of words in the training data that changed.** Shown are the number of word types and word tokens that changed, compared to the total number of word types and tokens on the source side of the different training bi-texts.

| | Bi-text(s) | Sentences | **Unknown** | | BLEU |
|---|---|---|---|---|---|
| | | | Types | Tokens | |
| 1 | ml-en | 160K | 3,115 | 7,101 | 17.90 |
| 2 | ml$_{in}$-en | 160K | 2,912 | 7,288 | 19.15 |
| 3 | in-en | 28.4K | 1,547 | 2,101 | 23.80 |
| 4 | in-en+ml-en | 28.4K+160K | 1,170 | 1,532 | 24.43 |
| 5 | in-en+ml$_{in}$-en | 28.4K+160K | 1,182 | 1,544 | 24.72 |

Table 14: **Unknown words for the Indonesian test dataset.** Shown are the number of unknown word types and word tokens, and Bleu score in % for different training bi-texts and simple bi-text concatenations (**cat×1**). The counts are with respect to the training bi-text; the actual number of unknown words at translation time can differ. Indonesian bi-texts were for tuning and testing and $en_{in}$ monolingual data was used for language modeling.

Next, we studied the impact of transliteration on the number of unknown words for the test data. Table 14 shows the results for Indonesian, using Indonesian bi-texts for tuning and testing and $en_{in}$ monolingual data for language modeling. Comparing lines 1 and 2, we can see that transliteration has a very limited impact on reducing the number of unknown word types when training on Malay data: the number of unknown word types drops only slightly from 3,115 to 2,912, while the number of unknown word tokens actually grows, from 7,101 to 7,288. Yet, there is an improvement in BLEU, from 17.90 to 19.15. This improvement is consistent for all $n$-gram scores included in BLEU: 1-gram (48.46 *vs.* 50.12), 2-gram (22.09 *vs.* 23.46), 3-gram (12.49 *vs.* 13.54), and 4-gram (7.67 *vs.* 8.44). Thus, apparently, the number of unknown word types is more important than the number of unknown word tokens. Moving down to line 3, we can see that the number of unknown word types is only halved when training on Indonesian instead of Malay, which confirms once again the similarity between Indonesian and Malay. Comparing lines 4 and 5, we can see that when we concatenate the Malay and Indonesian training bi-texts, the impact of transliteration is minimal: both in terms of word types/tokens and BLEU.

|    | Bi-text(s) | Sentences | Unknown Types | Unknown Tokens | Bleu |
|----|-----------|-----------|--------|--------|------|
| 1  | pt-en | 160K | 3,973 | 17,580 | 6.35 |
| 2  | pt$_{es}$-en | 160K | 1,574 | 10,337 | 11.35 |
| 3  | it-en | 160K | 5,529 | 23,088 | 4.06 |
| 4  | it$_{es}$-en | 160K | 2,413 | 13,492 | 9.38 |
| 5  | es-en | 160K | 362 | 440 | 27.90 |
| 6  | es-en+it-en | 160K+160K | 347 | 406 | 27.65 |
| 7  | es-en+it$_{es}$-en | 160K+160K | 316 | 374 | 27.69 |
| 8  | es-en+pt-en | 160K+160K | 273 | 295 | 27.83 |
| 9  | es-en+pt$_{es}$-en | 160K+160K | 240 | 257 | 28.14 |
| 10 | es-en+pt-en+it-en | 160K+160K+160K | 264 | 280 | 27.89 |
| 11 | es-en+pt$_{es}$-en+it$_{es}$-en | 160K+160K+160K | 221 | 232 | 28.02 |

Table 15: **Unknown words for the Spanish test dataset.** Shown are the number of unknown word types and word tokens, and Bleu score in % for different training bi-texts and simple bi-text concatenations (**cat×1**). The counts are with respect to the training bi-text; the actual number of unknown words at translation time can differ. Spanish bi-texts were used for tuning and testing.

The relative differences in the number of unknown words are much more sizeable when transliterating Portuguese/Italian to Spanish, as Table 15 shows. Comparing lines 1-2 and 3-4, we can see the number of unknown word types/tokens is about halved, while BLEU doubles. This confirms once again the importance of transliteration for these languages. Going down to line 5, we can see a 10-15 times drop in the number of unknown word types when training on the Spanish bi-text. This drop looks drastic compared to that for Malay in Table 14, but it can be partly explained by the larger size of the training *es-en* bi-text, which contains 160K sentence pairs compared to only 28.4K pairs for *in-en*. Since the *es-en* bi-text has reduced the number of unknown word types to 362, it becomes very hard to reduce this number further. Still, as lines 6-11 show, concatenating *es-en* with *pt-en* and *it-en* yields sizable improvements over using *es-en* only. Moreover, transliteration helps consistently in reducing the number of unknown words further and these reductions are bigger than those for Malay not only in relative, but also in absolute terms. Still, these reductions in the number of unknown words are not so great in absolute terms, and thus the corresponding differences in BLEU are small. Yet, transliteration yields consistent improvement for all concatenations: Spanish+Italian, Spanish+Portuguese, and Spanish+Portuguese+Italian.

Overall, we can conclude that large relative drops in the number of unknown words correspond to sizable improvements in BLEU; however, the results for small relative differences are less conclusive: they correspond to small fluctuation in BLEU for Portuguese and Italian but to somewhat larger differences for Malay. This could be a feature of the much lower token/type ratio for Malay, which is an agglutinative language, and thus quite rich in wordforms: as Table 13 shows, the token/type ratio is only about 50 for Malay, while it is about 307 and 345 for Portuguese and Italian, respectively.

### 8.3 Relative Improvement

Finally, we address the important question of how much "real" data our method saves.

Figure 3 compares graphically the improvements over the baseline using our method with 160K vs. 1.23M *pt-en* sentence pairs and transliteration for different number of original training *es-en* sentence pairs. We can see from this figure that, with 10K "real" training *es-en* sentence pairs, using 160K additional *pt-en* and our method yields a BLEU score that is comparable to what is achieved with 40K "real" *es-en* sentence pairs, *i.e.*, we cut the necessary "real" data by a factor of four. We can further see that using 1.23M *pt-en* sentence pairs improves this factor to five. Similarly, for 20K "real" *es-en* training sentence pairs, our method achieves a BLEU score that would require 3–3.5 times as much "real" training *es-en* data for the baseline system to match.

Figure 4 summarizes these statistics, showing how many times more "real" data would be needed for the baseline to match the performance of our method. We can see that we cut this by a factor of 1.5–4 and 2–5, when using 160K and 1.23M additional *pt-en* sentences.

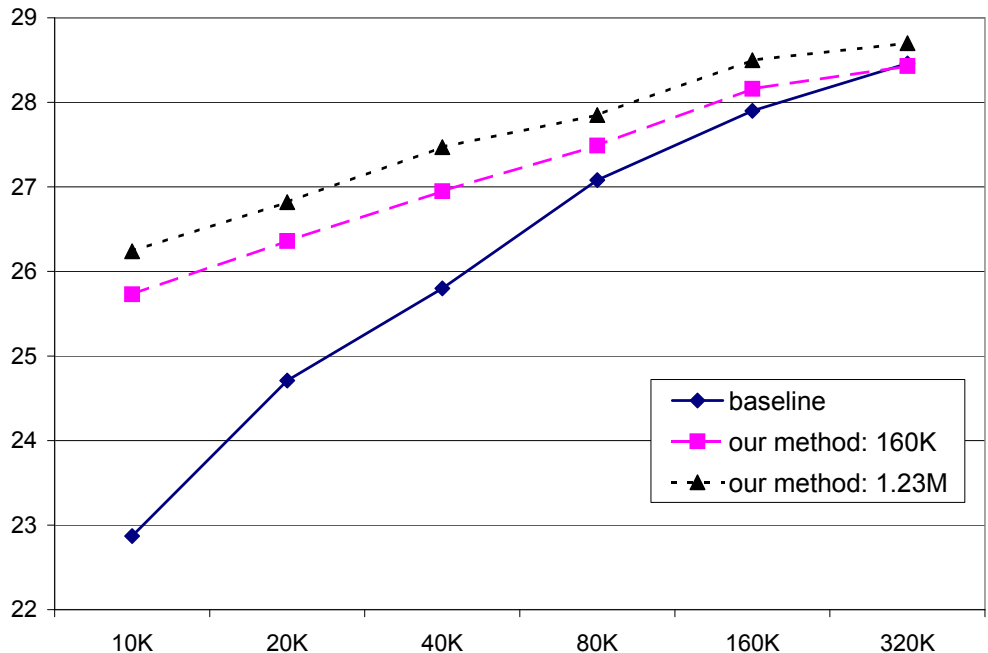

Figure 3: **Spanish→English: improvements over the baseline using our method with 160K vs. 1.23M *pt-en* sentence pairs and transliteration** for different number of original training *es-en* sentence pairs.

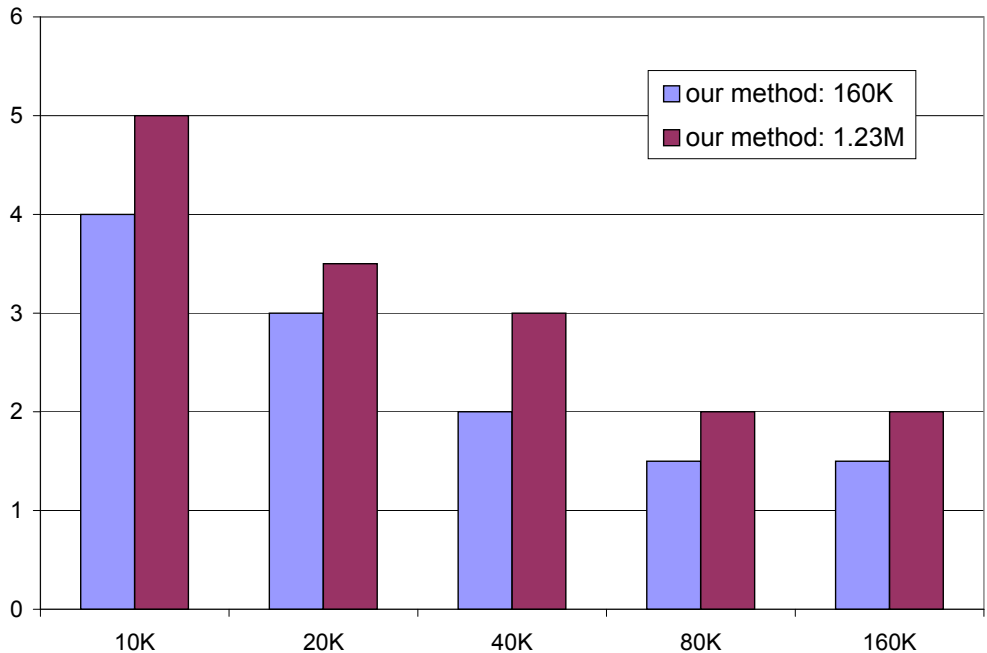

Figure 4: **Trade-off between Spanish→English and Portuguese→English data.**
Shown is the number of times we need to grow the original *es-en* training data in
order to achieve the same BLEU score as when using our method and 160K/1.23M
additional *pt-en* sentence pairs with transliteration.

## 9. Conclusion

We have proposed a novel language-independent method for improving statistical machine
translation for resource-poor languages by exploiting their similarity to related resource-rich
ones. We have achieved significant gains in BLEU, which improve over the best rivaling
approaches, while using much less additional data.

We further studied the impact of using a less closely related language as an auxiliary
language (Italian instead of Portuguese for improving Spanish→English SMT), we tried
using both Portuguese and Italian together as auxiliary languages, and we combined our
method with two orthogonal rivaling approaches: (1) using cognates between the source
and the target language, and (2) source-language side paraphrasing with a pivot language.
All these experiments yielded statistically significant improvements for small datasets.

Based on the experimental results, we can make several interesting conclusions:

1. We have shown that using related languages can help improve SMT: we achieved up to 1.35 and 3.37 improvement in BLEU for *in-en* (+*ml-en*) and *es-en* (+*pt-en*).

2. While simple concatenation can help, it is problematic when the additional sentences out-number the ones from the original bi-text.

3. Concatenation can work very well if the original bi-text is repeated enough times so that the additional bi-text does not dominate.

4. Merging phrase tables giving priority to the original bi-text and using additional features is a good strategy.

5. Part of the improvement when combining bi-texts is due to increased vocabulary coverage but another part comes from improved word alignments. The best results are achieved when these two sources are first isolated and then combined (our method).

6. Transliteration can help a lot in case of systematic spelling variations between the original and the additional source languages.

7. Overall, we reduce the amount of necessary "real" training data by a factor of 2–5.

In future work, we would like to extend our approach in several interesting directions. First, we want to make better use of multi-lingual parallel corpora, e.g., while we had access to a Spanish-Portuguese-English corpus, we used it as two separate bi-texts Spanish-English and Portuguese-English. Second, we would like to try using auxiliary languages that are related to the *target* language. Finally, we would like to experiment with more sophisticated ways to get the auxiliary language closer to the source that go beyond simple transliteration.

## Acknowledgments

We would like to thank the anonymous reviewers for their constructive comments and suggestions, which have helped us improve the quality of the manuscript. This research was supported by research grant POD0713875, and by the Singapore National Research Foundation under its International Research Centre @ Singapore Funding Initiative and administered by the IDM Programme Office.